\documentclass[sigconf]{acmart}
\usepackage{graphicx}
\graphicspath{ {./figures/} }
\usepackage{algorithm}
\usepackage{algpseudocode}
\usepackage{algorithmicx}
\usepackage{multirow}
\usepackage{algorithm,algpseudocode}
\usepackage{xcolor}
\usepackage{todonotes}
\usepackage{xcolor}
\usepackage{soul}
\usepackage{color}
\usepackage{balance}
\usepackage{amsmath}

\newcommand{\model}{\textbf{\textit{w}}}

\presetkeys%
    {todonotes}%
    {backgroundcolor=green!40}{}
\AtBeginDocument{%
  \providecommand\BibTeX{{%
    \normalfont B\kern-0.5em{\scshape i\kern-0.25em b}\kern-0.8em\TeX}}}

\renewcommand\footnotetextcopyrightpermission[1]{} 

\date{}

\begin{document}

\title{Let's Focus: Focused Backdoor Attack against Federated Transfer Learning}

\author{Marco Arazzi}
\affiliation{%
	\institution{University of Pavia}
	\streetaddress{}
	\city{Pavia}
	\state{}
	\country{Italy}
	\postcode{}
}
\email{marco.arazzi01@universitadipavia.it}

\author{Stefanos Koffas}
\affiliation{%
	\institution{Delft University of Technology}
	\streetaddress{}
	\city{Delft}
	\country{The Netherlands}}
\email{S.Koffas@tudelft.nl}

\author{Antonino Nocera}

\affiliation{%
	\institution{University of Pavia}
	\streetaddress{}
	\city{Pavia}
	\state{}
	\country{Italy}
	\postcode{}
}
\email{antonino.nocera@unipv.it}

\author{Stjepan Picek}
\affiliation{%
	\institution{Radboud University}
	\streetaddress{}
	\city{Nijmegen}
	\country{The Netherlands}}
\email{stjepan.picek@ru.nl}

\begin{abstract}
Federated Transfer Learning (FTL) is the most general variation of Federated Learning. According to this distributed paradigm, a feature learning pre-step is commonly carried out by only one party, typically the server, on publicly shared data. After that, the Federated Learning phase takes place to train a classifier collaboratively using the learned feature extractor. Each involved client contributes by locally training only the classification layers on a private training set. 
The peculiarity of an FTL scenario makes it hard to understand whether poisoning attacks can be developed to craft an effective backdoor. State-of-the-art attack strategies assume the possibility of shifting the model attention toward relevant features introduced by a forged trigger injected in the input data by some untrusted clients. Of course, this is not feasible in FTL, as the learned features are fixed once the server performs the pre-training step.

Consequently, in this paper, we investigate this intriguing Federated Learning scenario to identify and exploit a vulnerability obtained by combining eXplainable AI (XAI) and dataset distillation.
In particular, the proposed attack can be carried out by one of the clients during the Federated Learning phase of FTL by identifying the optimal local for the trigger through XAI and encapsulating compressed information of the backdoor class.
Due to its behavior, we refer to our approach as a focused backdoor approach (FB-FTL for short) and test its performance by explicitly referencing an image classification scenario.
With an average $80\%$ attack success rate, obtained results show the effectiveness of our attack also against existing defenses for Federated Learning.
    
\end{abstract}

\maketitle
\thispagestyle{empty}
\pagestyle{plain}

\section{Introduction}
\label{sec:introduction}

In recent years, deep learning and machine learning solutions received great attention from the research community thanks to the development of powerful technologies capable of handling and processing huge data volumes. However, the need to collect data from disparate sources to foster the training of models implies sharing raw data with a third-party computation unit, thus implying privacy concerns over the users' data.
This condition led to the development of Federated Learning~\cite{fl-mcmahan}, according to which users train collaboratively a model without sharing any private information and, therefore, maintaining data localized. Initially, the basic definition of Federated Learning assumes that all the involved clients own private data from the same feature space (Horizontal Federated Learning, HFL for short). This assumption makes this scenario impractical in some cases. As a matter of fact, in many real-life scenarios different parties need to use data for the same samples but from different feature spaces. For example, a bank may collaborate with an invoice agency to build a financial risk model for their customers~\cite{liu2024vertical}. Under this assumption, the collaborative learning among such clients is called Vertical Federated Learning (VFL). 

By extending the scenarios considered above, the most general case consists of parties that collaborate into a machine/deep learning process by using local data that differ in both feature and sample space; this situation is encompassed by the Federated Transfer Learning (FTL)~\cite{liu2024vertical}. 
Such a paradigm received great attention from the research community and has been, for instance, employed in~\cite{chen2020fedhealth}, where the authors propose a novel architecture for personalized healthcare via wearable devices. Given the type of application, we believe it is crucial to understand the vulnerabilities this paradigm introduces, which are still relatively unexplored. 
For this reason, this paper focuses on understanding whether backdoor attacks can be crafted by manipulating local data on maliciously controlled parties involved in Federated Transfer Learning.
Backdoor attacks have been largely studied in recent years in both HFL~\cite{badnets,bagdasaryan2020backdoor,wang2020attack,xie2019dba} and VFL~\cite{liu2020backdoor}; however, to the best of our knowledge, our proposal is the first to target FTL and show that such a paradigm is vulnerable to backdoor attacks. 
It is worth underlying that, compared to the classical FL scenarios, FTL presents unique challenges as the learning is typically broken down into two parts: a feature learning task, globally carried out by a server on a publicly shared dataset and the subsequent training of custom classification layers carried out by all the clients using private local data in a federated task. 
Because the feature extractor part of the classification model is learned in a different phase with respect to the federated learning one, crafting a backdoor implies a totally different approach than those typically adopted by existing backdoor attacks for FL.
In our study, to define our strategy for a backdoor attack, we adopt the FTL scenario of~\cite{chen2020fedhealth}, where the clients receive an initial model trained to extract features from input data by the server and then engage in a federated learning task to train only the classification layers (i.e., the first part of the model shared by the server is kept frozen). As typically done in the related literature, in our study, we chose image classification as a reference deep learning task, and the frozen part, pre-trained by the server, is the feature extractor of a Convolutional Neural Network (CNN). Then, each client uses a private local dataset to collaboratively fine-tune the last layers of the model (fully connected layers) by computing local updates during each training epoch and uploading the results to the server. The server then aggregates the changes and distributes the updated global model to the clients for the subsequent learning epochs. 
In this complex scenario, an adversary aiming at designing a backdoor trigger cannot leverage the possibility of forging new relevant features for the model to learn, such as a small square of noise~\cite{badnets} or simple objects like sunglasses~\cite{chen2017targeted}, because the feature extractor block of the model is frozen. Therefore, any additional feature will be promptly discarded by the global model. As a result, the attacker needs a more elaborated mechanism for the trigger design that would steer the poisoned sample's latent representation toward the target class. 

We overcome this challenge by combining an Explainable AI (XAI) step on the feature extractor with a data-driven approach to build a malicious trigger and suitably position it in high-attention locations learned by the global model. For this reason, we call our strategy a {\em focused} backdoor attack.
To do so, we leverage a powerful XAI tool, named GradCam~\cite{selvaraju2017grad}, to find out the best way for an adversary to alter the images. Moreover, we exploit a distillation strategy~\cite{zhao2020dataset} to encapsulate inside the trigger the main information characterizing the features of the target class of the backdoor. We argue that by suitably locating our trigger in attention locations for the feature extractor component of the model and encapsulating a compressed representation of the target class in it, we can deviate the behavior of the classifier when such a trigger is present in the input image.
The experiments we carried out show that FTL is still vulnerable to backdoor attacks built according to our strategy. 

In summary, our main contributions are as follows:
\begin{itemize}
    \item We introduce FB-FTL, the first focused backdoor attack against Federated Transfer Learning. Based on our experiments, FB-FTL achieves a success rate of over $80\%$, on average. It is important to note that, to our knowledge, this paper is the first work to study a backdoor attack against Federated Transfer Learning.
    \item We also extend the family of explainability-based attacks, defining and proving the importance of a ``focusing'' strategy for the positioning of the trigger using GradCam. In addition, we exploit dataset distillation to generate a trigger containing the features of the target class.
    \item We propose a strategy to blend better the pixels of the trigger with the colors of the original image, making it less noticeable.
    \item We tested our approach against well-known defenses for Horizontal Federated Learning, and we also compared it against a novel Vertical Federated Learning countermeasure based on Label Differential Privacy. While some defenses are effective with specific datasets, none of the considered defenses can mitigate our approach in all the considered scenarios.
\end{itemize}


An important facility of our proposal is the capability of adapting the trigger to the attacked data to minimize its impact in terms of necessary variations caused by its inclusion.
This is especially true for images where triggers typically alter visible regions to craft a backdoor.
Training data are automatically processed by local clients, and therefore, the fact that the trigger may actually be visible to some extent does not necessarily represent a blocking point for the attacker. However, minimizing its visible impact is crucial to make the attack as stealthy as possible. 


\section{Background}
\label{sec:background}
\subsection{Federated Learning}
\label{sub:federatedLeanirng}

Federated Learning (FL) allows $n$ clients to train a global model $\model$ collaboratively without revealing their datasets.  
According to~\cite{yang2019federated}, it can be classified into 1) Horizontal Federated Learning (HFL), 2) Vertical Federated Learning (VFL), and 3) Federated Transfer Learning.

In HFL, clients own data that share the same feature space but represent different entities~\cite{yang2019federated}. Each client trains a local copy of the model and then uploads its weights ($\{\model^i \mid i \in n\}$) to a server. HFL optimizes the following loss function:
\begin{equation}
    \mathop {\min }\limits_\model \ \ell (\model ) = \sum\limits_{i = 1}^n {\frac{{{k_i}}}{n}{L_i}(} \model ),\ \ {L_i}(\model ) = \frac{1}{{{k_i}}}\sum\limits_{j \in {P_i}} {{\ell _j}(\model, x_j)},
\end{equation}
where ${L_i}(\model )$ and $k_i$ represent the loss function and local data size of $i$-th client, and $P_i$ refers to the set of data indices with size $k_i$. At the $m$-th iteration, the training can be divided into three steps. First, all clients download the global model $\model_m$ from the server. Then, each client updates the local model by training with their datasets ($\model_m^i \leftarrow \model_m^i-\alpha\frac{\partial L(\model_m, b)}{\partial \model_m^i}$, where $\alpha$ and $b$ refer to learning rate and local batch). Finally, after the clients upload their local weights, the server updates the global model by aggregating the local weights. There are various aggregation techniques that can be used based on the given scenario. In this work, we will mostly consider averaging, but we will also test four additional aggregation strategies in an extra experiment.

In the VFL, clients own data that belong to the same entities but have different features~\cite{yang2019federated}. Thus, their local models can differ as their data use different features~\cite{arazzi2023blindsage}. 

\subsection{Federated Transfer Learning}
\label{sub:federatedTransferLerning}

In transfer learning, a model that has been trained for a specific task can be reused as a starting point for a model on a different task, significantly lowering the costs required for the development of the new model. Federated transfer learning (FTL) combines principles from Transfer Learning and Federated Learning. It consists of a central model that aggregates updates from clients or has trained a model from a publicly available dataset. This model is then distributed to the clients who use their own datasets to improve its performance on their data distribution~\cite{chen2020fedhealth}. Unlike Federated Learning, participants in FTL own datasets that have different population samples and features, making FTL a realistic scenario for real-world applications~\cite{jing2019quantifying}. In our case, we follow the FedHealth's paradigm~\cite{chen2020fedhealth}, where the clients freeze the first layers of the received CNN (convolutional and max pooling layers) and then train the fully-connected layers using their own private data.


\subsection{Backdoor Attacks}
\label{sub:backdoorAttacks}

The backdoor attack is a very popular threat against deep learning models introduced in~\cite{badnets}. In this attack, the adversary adds a secret functionality into a model that will be activated during inference. The backdoor is activated by malicious inputs that contain a predefined property, the trigger. The backdoor's activation allows the attacker to control the model's behavior. For example, in autonomous driving, the backdoored model could classify a stop sign as a speed limit~\cite{badnets}. The backdoor can be embedded through data poisoning~\cite{badnets}, code poisoning~\cite{blind-backdoors}, or model poisoning~\cite{handcrafted-backdoors-in-dnns}. To measure the attack's effectiveness, we use the attack success rate, which represents the number of times that the backdoor is activated over the total poisoned samples fed into the network. To keep the attack stealthy, we need to ensure that the model's performance on the designated task is not affected by the backdoor insertion.

\subsection{GradCam}
\label{sub:gradCam}

Interpretability in AI seeks to understand a model's decision. Various techniques can be used to this end. Recently, class activation mapping (CAM) was introduced~\cite{learning-deep-features-for-discriminative-localization}. CAM visualizes the areas of an image that are important for the model's decision but sacrifices the model's performance because it needs to modify the model's architecture~\cite{selvaraju2017grad}.
Gradient-weighted CAM (GradCam) is a generalization of CAM that does not have this requirement~\cite{selvaraju2017grad}. In particular, instead of modifying the model's architecture, it uses the gradient information that flows into the final convolutional layer to produce a localization of the crucial areas of the image that are connected to the model's decision. 
In this work, we use GradCam to identify the best position of our triggers.

\section{Methodology}
\label{sec:methodology}


\subsection{Attack Scenario}
\label{sub:attackScenario}

As introduced in Section~\ref{sub:federatedTransferLerning} and as also done in~\cite{chen2020fedhealth}, in the considered scenario, we assume that the server $S$ holds the public dataset $D_p=\{d_p, y_p\}$ and uses it to train the transfer model $M_p$. Without loss of generality, in our proposal and experiments, we consider an image classification task, making Convolutional Neural Networks an intuitive architectural choice. However, we argue that with minor tweaks, our approach could also work with different models.
In practice, the server carries out a global Feature Learning task that will then be shared by all the clients.
The weights $W_p$ of the model $M_p$ are trained as follows ($CE=$ cross-entropy):
\begin{gather}
    preds = M_p(d_p) \\
    l = CE(preds,\ y_p); \ \nabla W_p =\sum \frac{\partial l}{\partial M_p}. 
\end{gather} 

The trained model $M_p$ can now be propagated to the clients $C=\{c_1, \dots, c_n\}$.
In particular, each client knows the public dataset $D_p$ and holds a private dataset $D_i$ (where $i \in [1, n]$) that cannot be accessed by the others.
During the Federated Learning process, the lower levels of the model, in our case, the Convolutional layers $Conv$, are frozen to preserve the network's ability to extract low-level and more general features of the image.
The classification layers $fc$, instead, are the most capable of capturing the high-level features.
Therefore, the $fc$ layers are trained by the clients using their local datasets $D_i$ to customize the classification behavior based on the domain their private data are extracted from.
In particular, the training of $fc$ is obtained by optimizing the combination of two different loss functions.
The first one is a standard cross-entropy loss between the prediction on $D_i$ and the corresponding labels.
The second one is an alignment loss on the last layers of $fc$ to better personalize the model on the client's data. As described in~\cite{chen2020fedhealth}, this alignment loss function is intended to align the output features between the inputs.
In this particular case, the alignment is performed between the outputs of the model $M_p$ on the public dataset $D_p$ and the private datasets $D_i$. The loss function is calculated as follows:
\begin{gather}
    S_{img} = M_p(D_p); \ T_{img} = M_p(D_n) \\
    l_{CORAL} = \frac{1}{4d^2}|| \ S_{img} - T_{img} \ ||^2_F \\
    tot\_l = CE(M_p(D_i), y_i) + \alpha l_{CORAL}.
\end{gather}
Here, $||.||^2_F$ is the squared matrix Frobenius norm, $d$ is the dimension of the embedding features, and $\alpha$ is the trade-off parameter between the loss functions.
This alignment allows each client to obtain a personalized model $M_i$, which is more accurate for the local data.

Keeping the $Conv$ layers frozen, along with the availability of public data used to train the initial model, opens the framework to possible threats from a malicious client.
The intuition behind our approach is to exploit the transfer model to craft dynamic triggers that embed the features of the target class. Since the $Conv$ layers work as feature extractors in the considered scenario (image classification), they will focus only on the features of the images that are known from the pre-training of the network on the public data. For this reason, we can reasonably expect that the $Conv$ layers will discard traditional triggers that add additional features unknown to the network.
To craft our dynamic trigger, we need to distill low-level and general features leaked directly from $M_p$ and encapsulate them into it.
Since the transfer model $M_p$ and the public dataset $D_p$ are available for all the clients by design, an attacker can exploit them to craft triggers by leveraging features from $D_p$ that are known to the $Conv$ layers of the target model.

However, distilling a dynamic trigger is still not enough to drift the transfer model $M_p$ towards the target class, as we prove experimentally in Section~\ref{sub:gradcam_importance}. 
In this scenario, the positioning of the trigger plays a key role in the attack's success.
In our attack, we propose a strategy that aims to detect and override the main features of the victim image with a dynamic trigger containing compressed features of the target class, as described in Section~\ref{sub:attackName}.
Empowered by the generated triggers, we hence assume that the attacker controls a percentage of the clients to poison the model during the Federated Transfer Learning process.

\begin{figure*}[!ht]
    \centering
    \includegraphics[width=0.75\textwidth]{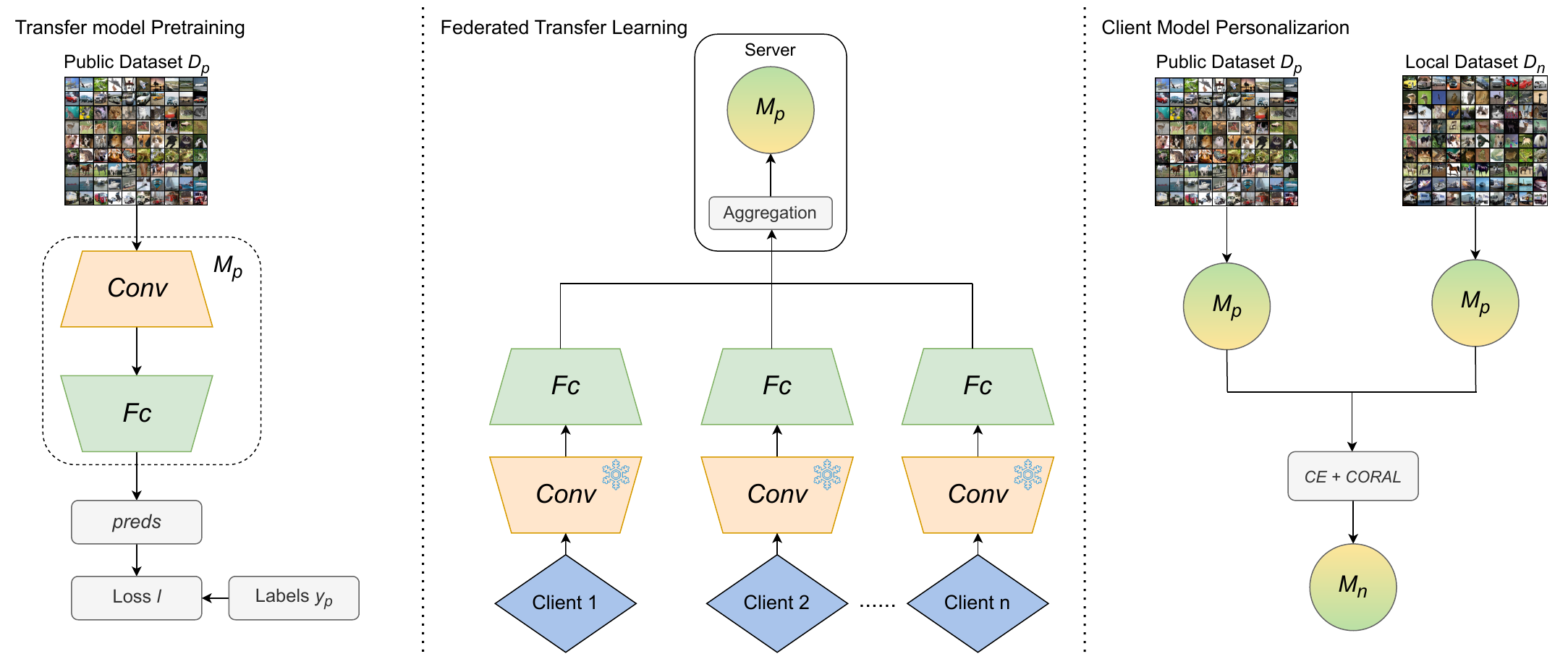}
    \caption{Federated Transfer Learning Framework.}
    \label{fig:FTLFramwork}
\end{figure*}

\subsection{Threat Model}
\label{sub:threatModel}

\textbf{Attacker Knowledge:} the adversary has access to the public dataset $D_p$, the public model $M_p$ provided by the server, and a private dataset $D_i$. Additionally, the adversary has access to the gradients and the feature maps of the model's layers.\\
\textbf{Attacker Capabilities:} using the private dataset $D_i$, the adversary is able to fine-tune the public model locally. However, the model's shallow layers, i.e., the convolutional and the max-pooling layers, are frozen, so only the fully connected layers can be altered. \\
\textbf{Attacker Goal:} by altering samples from the private dataset $D_i$, the attacker aims to inject a backdoor in the local model $M_i$. This backdoor cannot depend on the frozen layers, so a sophisticated trigger design is needed. After the backdoor is injected into $M_i$, the server will aggregate the local updates from the clients and update the unfrozen layers of the public model $M_p$. This updated model will be sent back to the clients for further fine-tuning. The adversary's end goal is, hence, to embed a backdoor that remains active after the completion of the Federated Transfer Learning.
The backdoor can then be used to control the model's behavior through malicious inputs containing the trigger and force it to misclassify them to the target class.


\subsection{FB-FTL Attack}
\label{sub:attackName}

\begin{figure}[!ht]
    \centering
    \includegraphics[width=0.62\columnwidth]{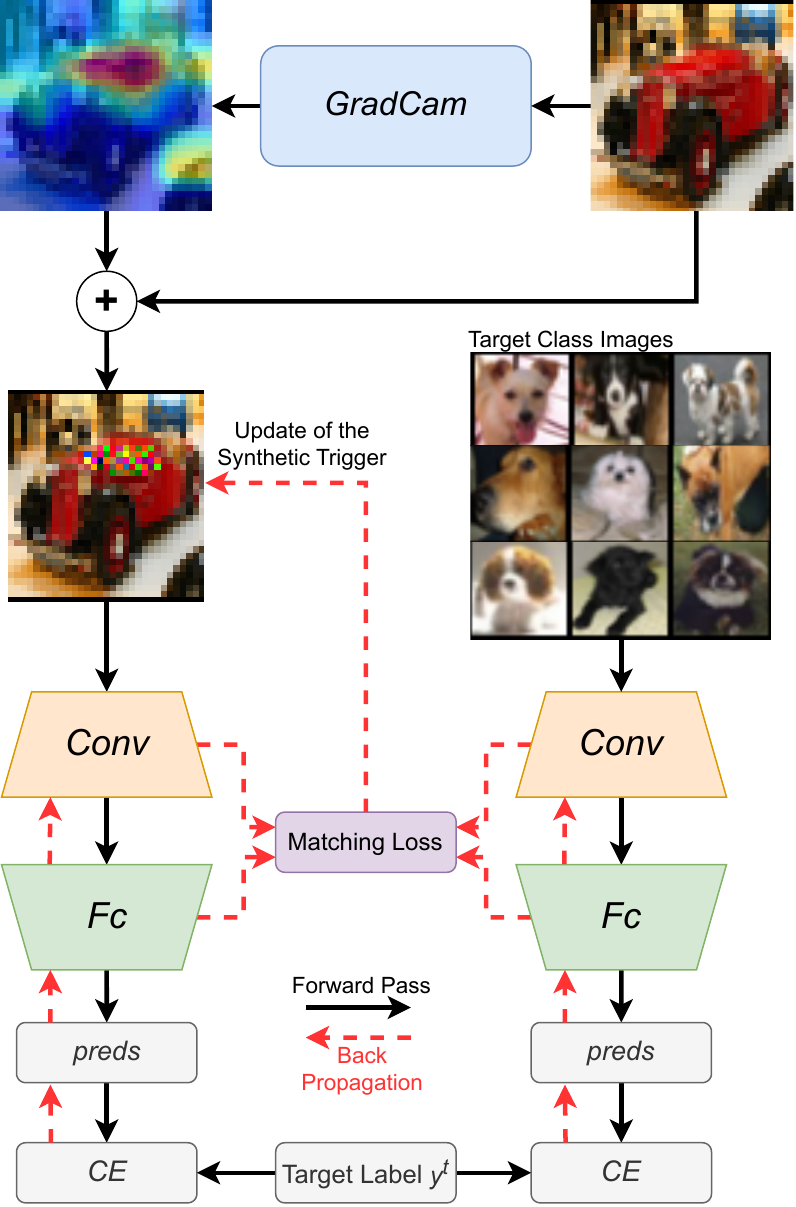}
    \caption{Trigger distillation.}
    \label{fig:triggerDistillation}
\end{figure}

As discussed in Section~\ref{sub:attackScenario}, the frozen convolutional layers $Conv$ are capable of extracting general low-level features from images that can be inserted as triggers in a victim image to perform a backdoor attack.
Our approach, shown in Figure~\ref{fig:triggerDistillation}, does not just add a trigger in a fixed position into the image. Instead, the idea behind this attack is to override the features of the original class with the ones of the target class.
To do so, an XAI approach that highlights the important regions in a target image for the given model is needed.
In particular, for this intent, we use GradCam~\cite{selvaraju2017grad}.
GradCam obtains the class-discriminative localization map for any class $c$ by computing the gradients for the given score of the class in the output $y^c$ concerning the feature maps $A^k$ of a convolutional layer.
The obtained gradients are then global-average pooled to obtain the neuron importance weights $a^c_k$ as follows:
\begin{gather}
    a^c_k = \frac{1}{Z} \sum_i \sum_j \frac{\partial y^c}{\partial A^k_{ij}}.
\end{gather}

These weights represent the importance of the features in the feature map for a target class $c$.
After this, a weighted combination of forward activation maps is performed and sent in input to a $ReLU$ activation function to obtain a heatmap of importance $I_m$ with the same shape as the feature maps: 
\begin{gather}
    I_m = ReLU \left( \sum_k a^c_kA^k \right).
\end{gather} 

The $ReLU$ activation function is applied to the linear combination of maps because we only consider the features with a positive impact on the class of interest. Negative pixels, instead, are likely to belong to other classes in the image~\cite{selvaraju2017grad}.
Without this $ReLU$, localization maps could emphasize more than just the desired class region, causing a lower localization performance.
The obtained map $I_m$ can now be used as a mask to select the regions of the image where the trigger should be injected.
In particular, we select the regions of the images where the weights of importance $a^c_k$ are higher than a given threshold $\tau$. 
Since the weights of importance $a^c_k$ fall in the interval of values $[0,1]$, $\tau$ can be picked in the same interval.
To inject a trigger capable of overriding the features of the original image with the main features of the target class, we use the intuition of the dataset distillation~\cite{zhao2020dataset}.
Specifically, the synthetic trigger is distilled using the Dataset Condensation technique with gradient matching. 
We take a subset of images $B_i$ for which we generate the GradCam heatmaps of the same size as the images.
Then, we set to $0$ the regions of the heatmaps lower than the given threshold $\tau$ and to $1$ the zones with a higher value, thus generating a mask.
The obtained mask can now be used to inject a trainable synthetic trigger initialized as random noise into the best location of the target images.
Our approach proceeds by exploiting the pre-trained transfer model $M_p$, for which we keep the parameter frozen for the entire process to learn the correct trigger.
At each iteration, we feed at first the images $B_i$, and we obtain the gradients $\nabla W^B_p$ on the model in relation to target class $y^t$ using the cross-entropy loss.
In a second step, instead, we repeat the process with a set of true images $T_i$ of the target class from the public dataset used to pre-train the model $M_p$. In the same way, we generate the related gradients $\nabla W^T_p$ on the parameters of $M_p$.
The idea is to inject into the trigger the generalized features from the data of the target class used to train the transfer model.
Through a distance metric, in our case, the cosine similarity, we calculate the loss on the distance between the generated gradients $\nabla W^B_p$ and $\nabla W^T_p$, and we backpropagate on the trainable trigger injected in $B_i$ as follows:
\begin{gather}
    \nabla W^B_p \leftarrow CE(M_p(B_i), y^t); \nabla W^T_p \leftarrow CE(M_p(T_i), y^t) \\
    B_i[I_m>\tau] \leftarrow COS(\nabla W^B_p, \nabla W^T_p).
\end{gather}
The process is then repeated for multiple iterations.

\subsubsection{Blended Trigger}
\label{sub:blendedTrigger}

The intuition behind our approach is, hence, to override the main features of an image, allowing its correct classification with a distilled trigger.
This strategy would inevitably alter the original image and, most probably, the main visual content, as shown in Figure~\ref{fig:examplesTriggers}.
This alteration can be detected using similarity metrics that compare the original image with its altered version.
To synthesize a trigger that blends better inside the image making it less noticeable to automatic detection systems or to a human eye, we can include a term, derived from a similarity metrics capable of detecting such alterations, in the distillation matching loss.

Concerning such similarity metrics, one well-fitting option can be the Learned Perceptual Image Patch Similarity $LPIPS$~\cite{zhang2018unreasonable}.
We follow the idea of using the $LPIPS$ metric from the Style Transfer and Generative Images approaches~\cite{liu2021content, jo2020investigating}.
These approaches employ a $LPIPS$ loss due to its capability of providing better feature space than other similar solutions for improving the perceptual quality of the resulting image.
In detail, using a pre-defined network like VGG or Alexnet, $LPIPS$ compares the activations of two image patches to compute the similarity. This measure has proven to match also the real perception of humans~\cite{zhang2018unreasonable}. 
A low LPIPS score means that the two images are perceptually similar.
With that said, our trigger distillation loss can be changed as follows:
\begin{equation}
    B_i[I_m>\tau] \leftarrow COS(\nabla W^B_p, \nabla W^T_p) + \lambda LPIPS(B_i,O_i),
\end{equation}
where $\lambda$ is the weight to module the contribution of $LPIPS$ on the overall loss, and $O_i$ is the original image without the trigger.

In the standard implementation of our approach, the pixels that compose the trigger are reinitialized as random Gaussian noise inside the original image $O_i$. A possible variation to improve the $LPIPS$ value even further is to back-propagate the combination of the matching loss and $LPIPS$ to the original pixel of the images without masking with random noise.
In this case, the process changes as follows:

\begin{gather}
    BO_i[I_m>\tau] \leftarrow COS(\nabla W^B_p, \nabla W^T_p) + \lambda LPIPS(BO_i,O_i),
\end{gather}
where $BO_i$ is the backdoored image with the trigger initialized with the same values as the original one.
As we can see, the approach is the same, except for the initialization of the trigger's pixels.

In Section~\ref{sub:lpips}, we present the comparison of our approach performance according to the considered combination of losses and initialization strategies of the trigger.

\section{Experimental Results}
\label{sec:Experiments}

\subsection{Experimental Setup}
\label{sub:experimentalSetup}


In our experiments, we focused on a testbed where we considered a Federated transfer learning framework composed of one server and multiple clients. In particular, in the baseline experiments, the number of clients is set to $10$, and the percentage of attackers is set by default to $30\%$.
We present additional results by changing the number of clients, in particular $\{5, 10, 100\}$
The considered main task is image classification. The considered baseline transfer model used as the feature extractor is a \emph{Resnet-18}~\cite{he2016deep}, kept frozen during the transfer learning process, with a $MLP$ classifier on top.
To showcase the generalizability of our approach, we also tested it with two additional architectures: ConvNet~\cite{gidaris2018dynamic} and VGG16~\cite{simonyan2014very}.
The different architectures according to the dataset are reported in Table~\ref{tab:modelsDetails}. 
The split of the datasets between train and test sets is left on default if available; otherwise, we considered $80\%$ of the data for the training and $20\%$ for testing. From the training set, $50\%$ is kept as public data to pre-train the model and the remaining half is split across the clients to be used as a private set. 
The pre-training of the model has been conducted using the public data for $10$ epochs with a learning rate of $0.1$ for the first five epochs and $0.01$ for the remaining five.
From our experimental campaign, this training strategy has proven to train the model with the partial dataset as the design of the framework and get the best accuracy.
The model is trained from scratch instead of restoring the ImageNet weights to be consistent with the scenario in which the model must be pre-trained only on public data~\cite{chen2020fedhealth}.
During the Federated Learning process, the classification layers are trained with a learning rate of $0.1$. 
The matching loss employed to distill the trigger is the \emph{Cosine Similarity} as presented in Section~\ref{sub:attackName}.

In the following sections, we present the attack's accuracy with the best value of importance threshold $\tau$ chosen between the set of possible values, specifically $\{0.5, 0.6, 0.7, 0.8, 0.9\}$.
We compare our solution with well-established backdoor attacks for HFL to understand if they are still effective in this configuration. In particular, we compared our solution with three static patch backdoor attacks: pixel pattern trigger, square trigger, and watermark trigger. In addition, we compared our solution against a novel approach that uses a trainable dynamic trigger: A3FL~\cite{NEURIPS2023_A3FL}. 

Moreover, we tested our solution with different percentages of attackers between the clients ($\{10\%, 20\%, 30\%, 40\%, 50\%\}$) to see if the effectiveness of our attack relies on the number of malicious clients. 
In Section~\ref{sub:gradcam_importance}, we will test the importance of the use of GradCam to spot the right position of the trigger compared to a squared dynamic trigger distilled in the same way as ours but with fixed coordinates in the corner of the image.

In Section~\ref{sub:lpips}, we redefine the learning loss of the dynamic trigger, combining with the original matching loss a similarity loss that preserves better the characteristics of the original image, making the trigger less noticeable but at the same time preserving most of the backdoor attack accuracy. In particular, we selected the $LPIPS$ loss function that measures the perceptual difference between the original images and their manipulated counterparts with the injected triggers.

For our experiments, we used a machine with an AMD Ryzen 5800X CPU paired with 32GB of RAM and an RTX 3070ti with 8GB of VRAM. Our attack has been developed using PyTorch 2.0.

\begin{table}[h]
\centering
\caption{Baseline model architectures. $MLP-N$ refers to Multi-Layer Perceptron with $N$ layers}
\label{tab:modelsDetails}
\resizebox{0.6\columnwidth}{!}{%
\begin{tabular}{ccc}
\hline
Dataset & \begin{tabular}[c]{@{}c@{}}{\em Conv} \\ Feature Extractor\end{tabular} & \begin{tabular}[c]{@{}c@{}}{\em Fc}\\ Classifier\end{tabular} \\ \hline
CIFAR10 & Resnet-18                                                         & MLP-4                                                   \\
CINIC10 & Resnet-18                                                         & MLP-4                                                   \\
SVHN    & Resnet-18                                                         & MLP-4                                                   \\
GTSRB   & Resnet-18                                                         & MLP-2                                                   \\ \hline
\end{tabular}
}
\end{table}

\subsection{Performance Evaluation of FB-FTL}
\label{sub:perfromance_evaluation}

In this section, we present the main results in terms of the success rate of our approach compared to existing solutions.
With this experiment, we want to investigate the applicability of backdoor attacks to this FTL.
Since no other attack in this setup is available, we decided to compare our solution against attacks that are intended for HFL due to their similarities in terms of the configuration of the training between server and clients. In both cases, the clients train the global model on their local data while the server aggregates compared to VFL, in which the global model is split between the top model on the server and the bottom models on the clients. 
In addition to traditional static triggers, we compare our approach also against a novel solution, named A3FL~\cite{NEURIPS2023_A3FL}, that injects in the image a dynamic trigger specifically generated for the target class.
Since one of FTL's main properties is freezing the feature extractor layers, we expect that traditional backdoor attacks will be ineffective in this scenario.
The considered attacks introduce into the image the triggers with additional features that are unknown to the feature extractor network.
Considering the A3FL solution, we expect better performance than the traditional static backdoors. However, we are still unable to match the performance of our attack due to the absence of the focus logic with GradCam.

\begin{table*}[!ht]
\centering
\caption{Main results}
\label{tab:mainResults}
\resizebox{0.95\textwidth}{!}{%
\begin{tabular}{cccccccccccc}
\hline
\multirow{2}{*}{\begin{tabular}[c]{@{}c@{}}Backdoor\\ Attacks\end{tabular}} & \multicolumn{2}{c}{CIFAR10}                                                      & \phantom{a} & \multicolumn{2}{c}{CINIC10}                                                      & \phantom{a} & \multicolumn{2}{c}{SVHN}                                                         & \phantom{a} & \multicolumn{2}{c}{GTSRB}                                                        \\ \cline{2-3} \cline{5-6} \cline{8-9} \cline{11-12} 
                                                                            & Model Accuracy & \begin{tabular}[c]{@{}c@{}}Backdoor\\ Success Rate\end{tabular} & & Model Accuracy & \begin{tabular}[c]{@{}c@{}}Backdoor\\ Success Rate\end{tabular} & & Model Accuracy & \begin{tabular}[c]{@{}c@{}}Backdoor\\ Success Rate\end{tabular} & & Model Accuracy & \begin{tabular}[c]{@{}c@{}}Backdoor\\ Success Rate\end{tabular} \\ \hline
No Attack                                                                   & 81.8\%         & -                                                               & & 71.0\%         & -                                                               & & 90.0\%         & -                                                               & & 95.0\%         & -                                                               \\
Pattern Trigger                                                             & 77.9\%         & 13.3\%                                                          & & 67.4\%         & 13.5\%                                                          & & 87.4\%         & 8.7\%                                                           & & 93.4\%         & 4.7\%                                                           \\
Square Trigger                                                              & 75.4\%         & 17.6\%                                                          & & 67.7\%         & 11.6\%                                                          & & 87.3\%         & 7.8\%                                                           & & 92.6\%         & 5.4\%                                                           \\
Watermark Trigger                                                           & 76.5\%         & 19.3\%                                                          & & 67.8\%         & 12.9\%                                                          & & 88.3\%         & 9.7\%                                                           & & 93.9\%         & 4.1\%                                                           \\
\hline
A3FL~\cite{NEURIPS2023_A3FL}                                                & 73.1\%         & 31.2\%                                                          & & 65.5\%         & 24.3\%                                                          & & 87.0\%         & 10.0\%                                                          & & 94.0\%         & 10.1\%                                                          \\
FB-FTL (our)                                                          & 81.3\%         & 91.1\%                                                          & & 70.2\%         & 73.3\%                                                          & & 89.6\%         & 72.1\%                                                          & & 94.7\%         & 86.5\%                                                          \\ \hline
\end{tabular}%
}
\end{table*}

From Table~\ref{tab:mainResults}, as expected, the existing attacks are almost unperceived by the federated model.
The A3FL approach, instead, achieves a better success rate than static triggers while still affecting the model accuracy on the main task like the traditional one.
These results also give some insights into the importance of focusing logic with GradCam.
In Section~\ref{sub:gradcam_importance}, we present an in-depth analysis in this sense.
Looking at our approach FB-FTL, we can see it is effective across the considered datasets. 
It is interesting to observe how our approach is capable of achieving high success rates despite the different characteristics of the datasets presented in Appendix~\ref{sub:datasets}. 
Even with datasets like SVHN, where the background contains information about other classes, our attack distills a trigger that preserves the importance of the original features compared to the ones in the background.
Considering GTSRB, instead, since contains images of traffic signs, the samples shares many features and in this case, our attack distills into the trigger the right ones selecting the features belonging just to the target class.
We can see also how our approach is the best at being effective and, at the same time, only slightly affects the performance of the model on the main task. 
This is because the feature extractor part of the model will recognize only the known features, so with traditional backdoor attacks, the network will still focus just on the information of the original class, discarding the trigger. Basically, the attacker is training the model, assigning to these images the target class, drifting the accuracy of the model on those features, affecting the final accuracy. 
In our case, instead, we are substituting the original features with information from the target class distilled from the network itself. In this way, the network will recognize the new features as known, only slightly affecting the accuracy of the main model.

\begin{figure}[!ht]
    \centering
    \includegraphics[width=0.3\textwidth]{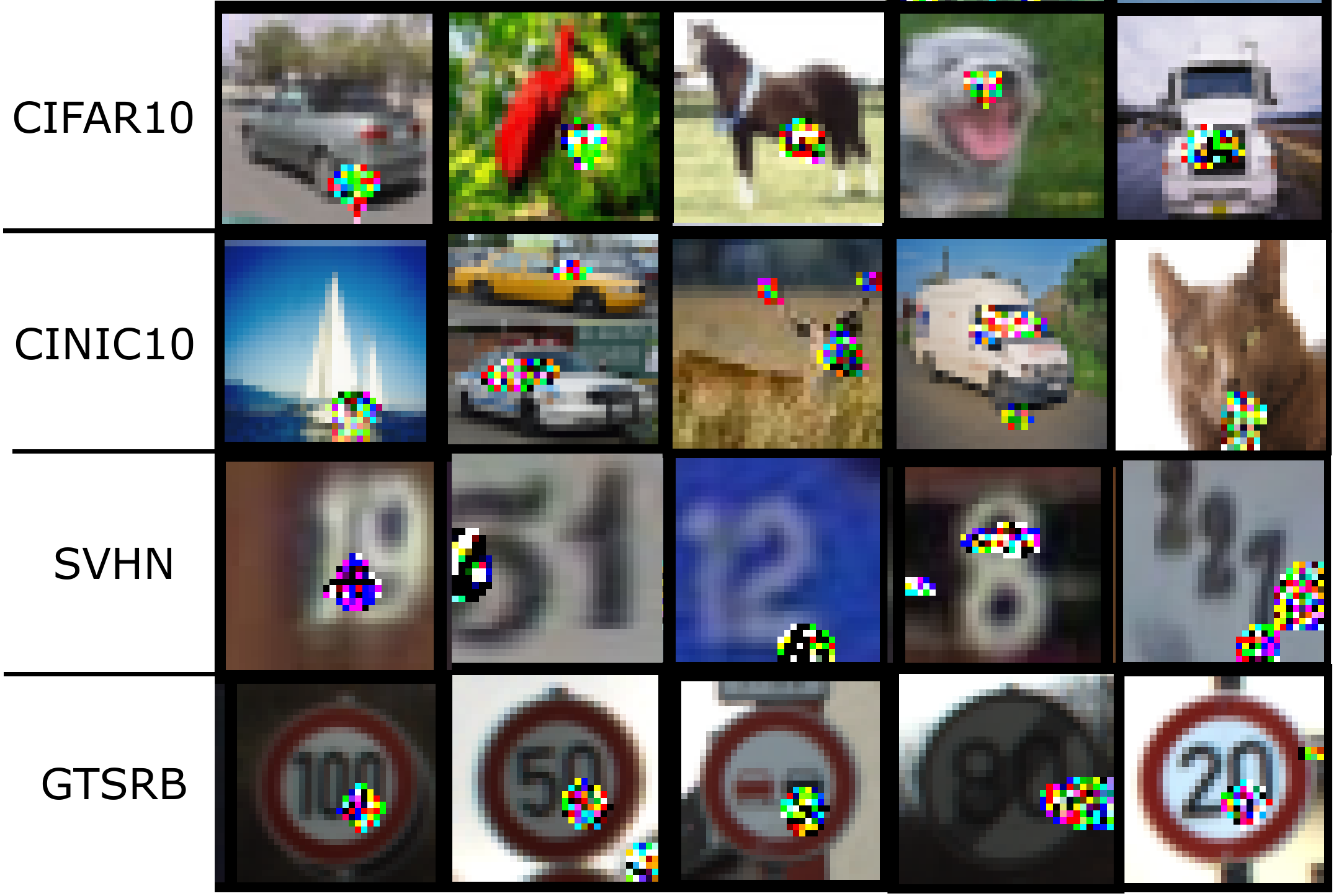}
    \caption{Examples of distilled triggers}
    \label{fig:examplesTriggers}
\end{figure}

In Figure~\ref{fig:examplesTriggers} and Appendix~\ref{sec:moreExamples}, we show some examples of triggers generated for different classes across the considered datasets.
The position of the trigger changes according to the region of importance spotted by the GradCam algorithm.
In Section~\ref{sub:gradcam_importance}, we present an ablation study that demonstrates the importance of finding the best region of the image where to distill the trigger instead of having fixed coordinates.
On the other hand, this approach also owes its success to the overriding of the main features of the image using GradCam to position the trigger.
This makes inevitable, in most of the images, an alteration of the main subject of the image.
In Section~\ref{sub:lpips}, we present the result of our approach, adding the contribution of the $LPIPS$ metric in the loss calculation to make the distilled trigger less noticeable to the human eye.

\subsection{Results Changing Percentage of Attackers}
\label{sub:numAttackers}

In this section, we assess how the performance of our attack changes according to the percentage of attackers between the clients. 
As we stated in Section~\ref{sub:experimentalSetup}, the previous results were collected by setting the percentage of attackers equal to $30\%$.
In this experiment, due to the nature of our attack, which relies mainly on the frozen $Conv$ layers of the model, we expect that the success rate of the attack would not drastically change from the baseline reported in the previous section. 
Thus, we aim to confirm our intuition that the success of our attack is only partially correlated to the number of attackers that poison the local updates.

\begin{figure}[!ht]
    \centering
    \includegraphics[width=0.85\columnwidth]{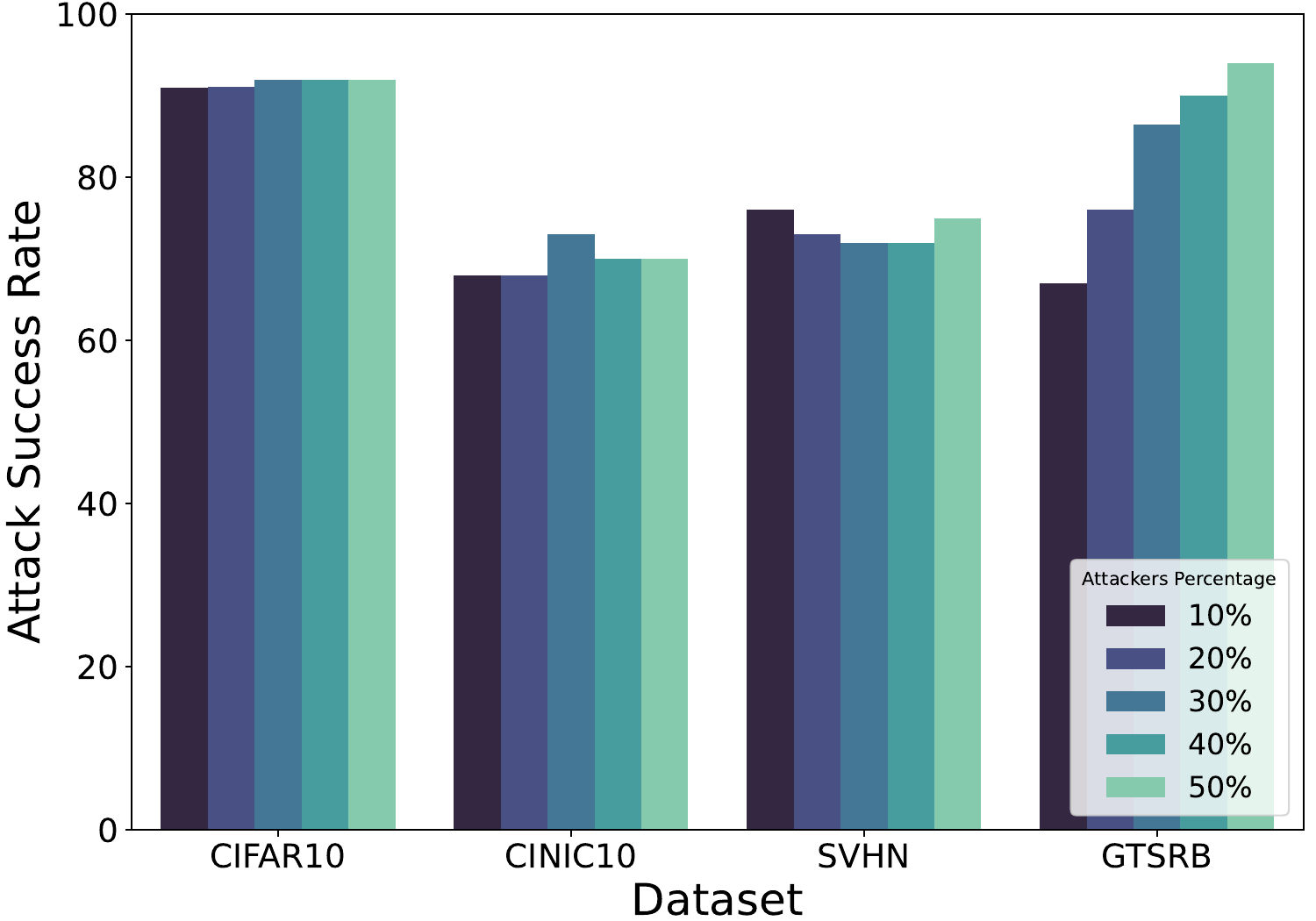}
    \caption{Performance of the attack changing the percentage of attackers.}
    \label{fig:attackersPerc}
\end{figure}

In Figure~\ref{fig:attackersPerc}, we show the results of this experiment.
As we can see, with three out of four datasets, we can confirm our intuitions. 
Indeed, the success rate of the attack is comparable across the different scenarios with negligible fluctuations.
The only dataset that benefits from a high percentage of attackers is GTSRB.
With the percentage of attackers of $50\%$, the success rate is close to $100\%$.
If we look, instead, at the worst scenario, the accuracy of the attack is a little lower than the baseline but still with a not negligible success rate of around $60\%$.
This is expected due to the similarities between classes of the dataset requiring for the attacker to contribute to the training of the federated model with the backdoored images to make the attack more effective.

The previous experiment was conducted with a total number of number of clients equal to $10$.
In this sense, we conducted a second experiment varying the number of clients included in the federated process.
In particular, we want to verify how our attack varies, considering just one malicious client but changing, in this case, the number of clients.
In the first scenario, we considered half of the clients compared to the baseline, $5$ clients in total, and the second scenario includes ten times more clients, specifically $100$.
Since our attack relies on the characteristic that the feature extractor in the considered Federated Transfer Learning scenario is frozen, we expect that our attack is just partially dependent on the total number of clients in the Federated Learning process.

\begin{figure}[!ht]
    \centering
    \includegraphics[width=0.85\columnwidth]{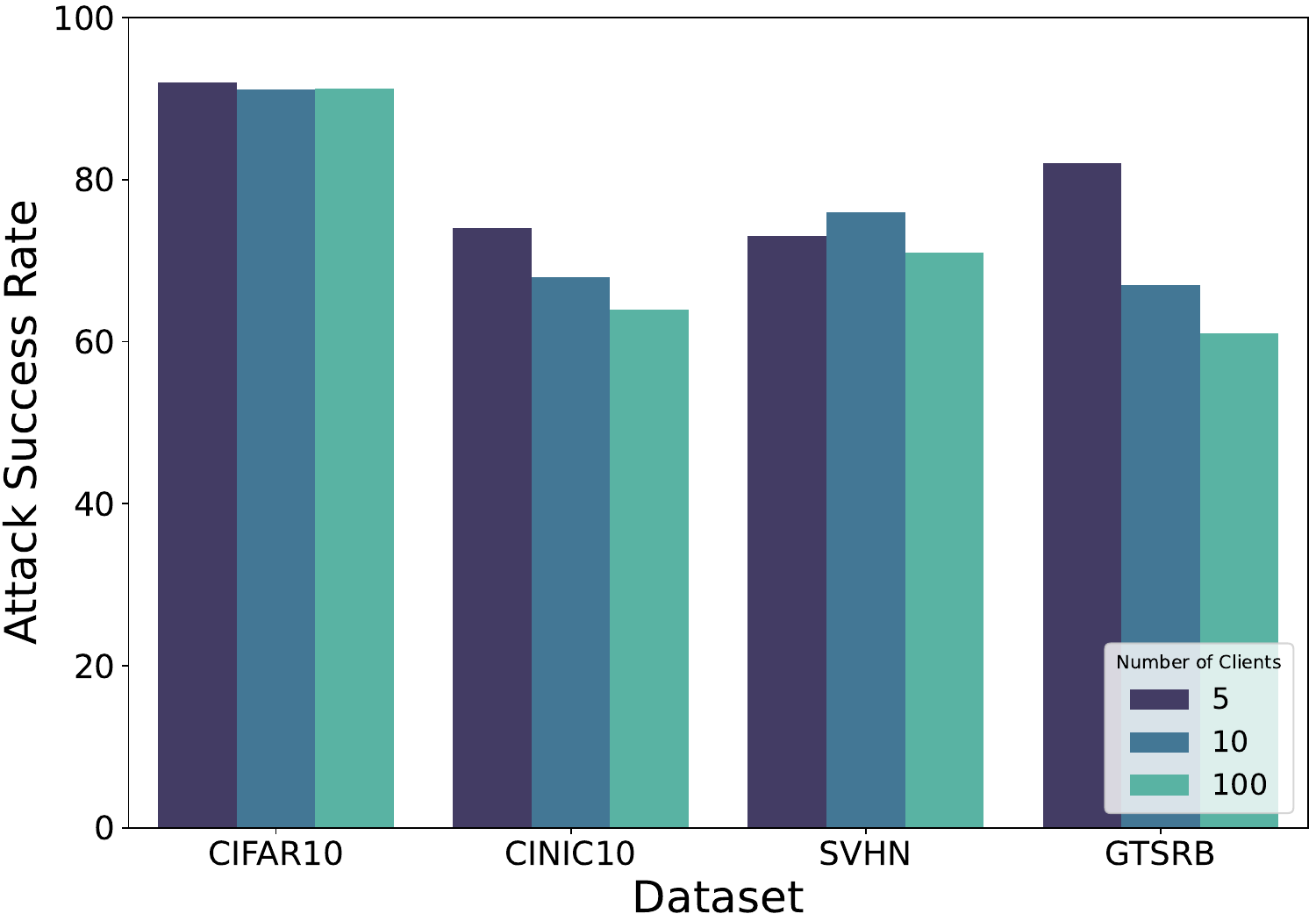}
    \caption{Performance of the attack changing the total number of clients.}
    \label{fig:attackersNumClientss}
\end{figure}

The results of this experiment are shown in Figure~\ref{fig:attackersNumClientss}.
As anticipated, we can notice the difference between $10$ and $100$ is just $5\%$ at maximum, preserving a success rate over $60\%$ across all the datasets.
Instead, looking at the scenario with $5\%$, we can see how the performance is better than considering $10\%$ as expected, but similarly to the previous use case, the difference is almost negligible.
The highest recorded difference is about $15\%$ for the GTSRB dataset with $5$ total clients compared to the baseline of $10$.
Even if the proportion between malicious and benign clients is doubled with $5$ overall clients compared to $10$ considering just one attacking client, the success rate of our attack is not two times higher with double the importance of the malicious client contribution in the averaging of the updates with $5$ total clients.
These results allow us to confirm that our solution is independent of the percentage of attackers between the parties and the total number of clients participating in the federated training.

\subsection{Evaluation Changing $\tau$ Threshold}
\label{sub:analysis_on_tau}

In this section, we present the results of the experiments varying the $\tau$ threshold used to select the region of the trigger according to the weights of importance returned by GradCam as presented in Section~\ref{sub:attackName}.
In particular, changing this value will affect positively or negatively not only the performance of our attack but also the integrity of the original image.
Intuitively, if we set $\tau$ to a low value, we will inject a trigger that occupies a large portion of the image. Since the aim of our attack is to distill an effective trigger but at the same time preserve most of the content of the image, the $\tau$ value is a parameter that has to be tuned.
In particular, the idea is to select the highest value of $\tau$ that preserves the performance of the model while altering as little as possible the original image.

\begin{figure}[!ht]
    \centering
    \includegraphics[width=0.85\columnwidth]{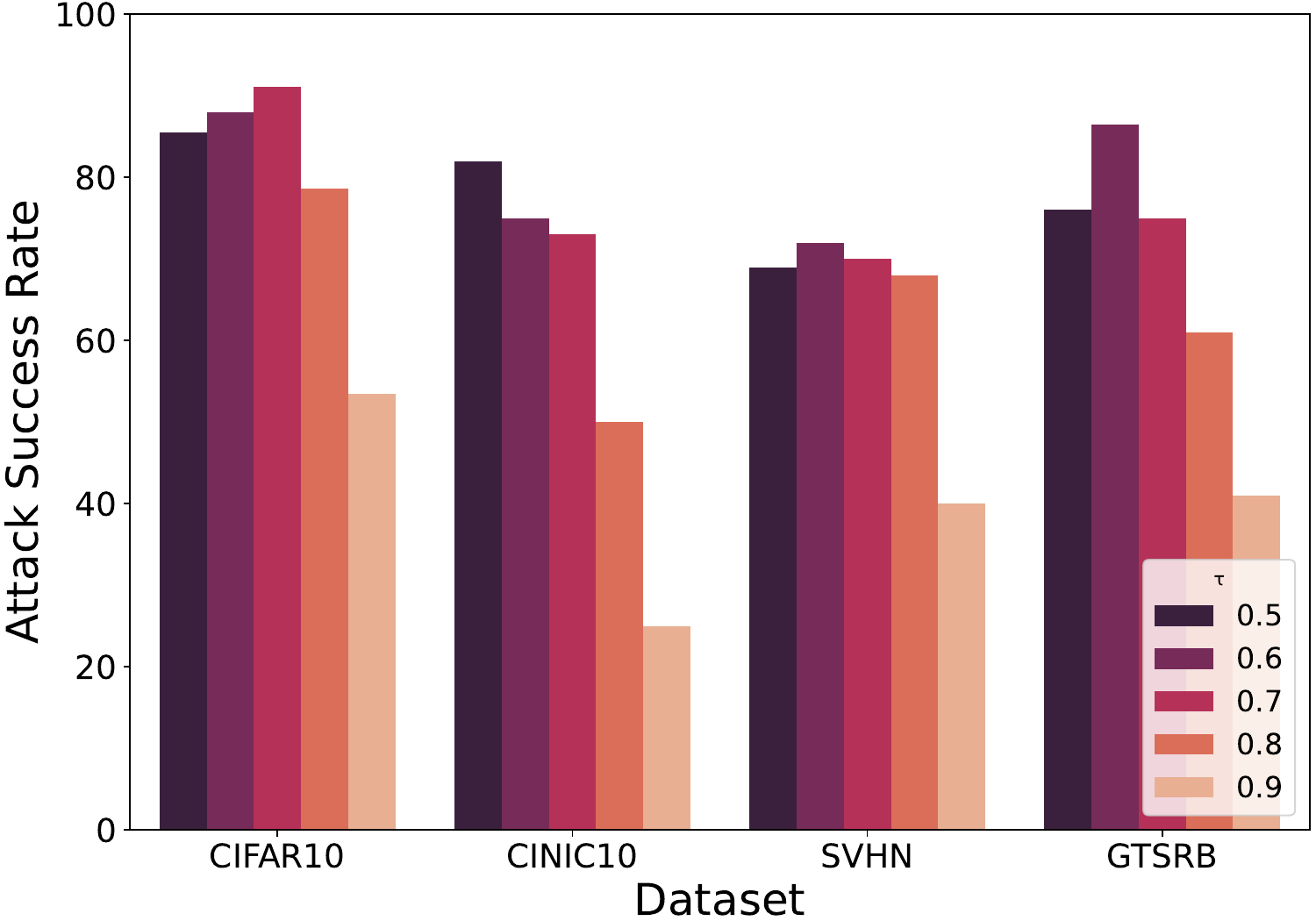}
    \caption{Backdoor success rate changing $\tau$ value.}
    \label{fig:importanceAnalysis}
\end{figure}

Results of this experiment are reported in Figure~\ref{fig:importanceAnalysis}.
As expected, with high values of $\tau$, e.g., $0.8$ or $0.9$, the performance of our approach drops.
This is because, with high values, the region dedicated to the trigger is too small to embed the features necessary to make our approach successful. 
At the same time, the region is not big enough to override the main features of the image.

In Section~\ref{sub:gradcam_importance}, we prove this by showing the importance of the positioning of the trigger to cover the main features of the original image.
Looking at lower values of $\tau$, instead, we can see how having a smaller $\tau$, and by consequence, a larger trigger, does not necessarily benefit the accuracy of the model.
In this situation, because there is a bigger region to distill, our attack has more space to include more refined features about the target class.
Since dataset distillation works by providing the network examples of the target class, having a bigger region to distill makes our trigger prone to overfit too much on example images.
In this way, the trigger will include information specific to those particular images in addition to the more general features of the class.
This additional information breaks our assumption of distilling triggers with only general features of the target category, making them less relevant and affecting the attack's performance.

\subsection{GradCam Importance Analysis}
\label{sub:gradcam_importance}

As presented in Section~\ref{sub:attackName}, GradCam is one of the main components of our approach. Dynamic triggers alone, as we showed in Table~\ref{tab:mainResults}, are not enough to produce backdoor attacks effective in the Federated Transfer Learning Scenario.
In this section, we present an analysis to assess the importance of an explainability technique, in our case, GradCam, to spot the most important features of the original images to be covered by our distilled trigger.

\begin{figure}[!ht]
    \centering
    \includegraphics[width=0.3\textwidth]{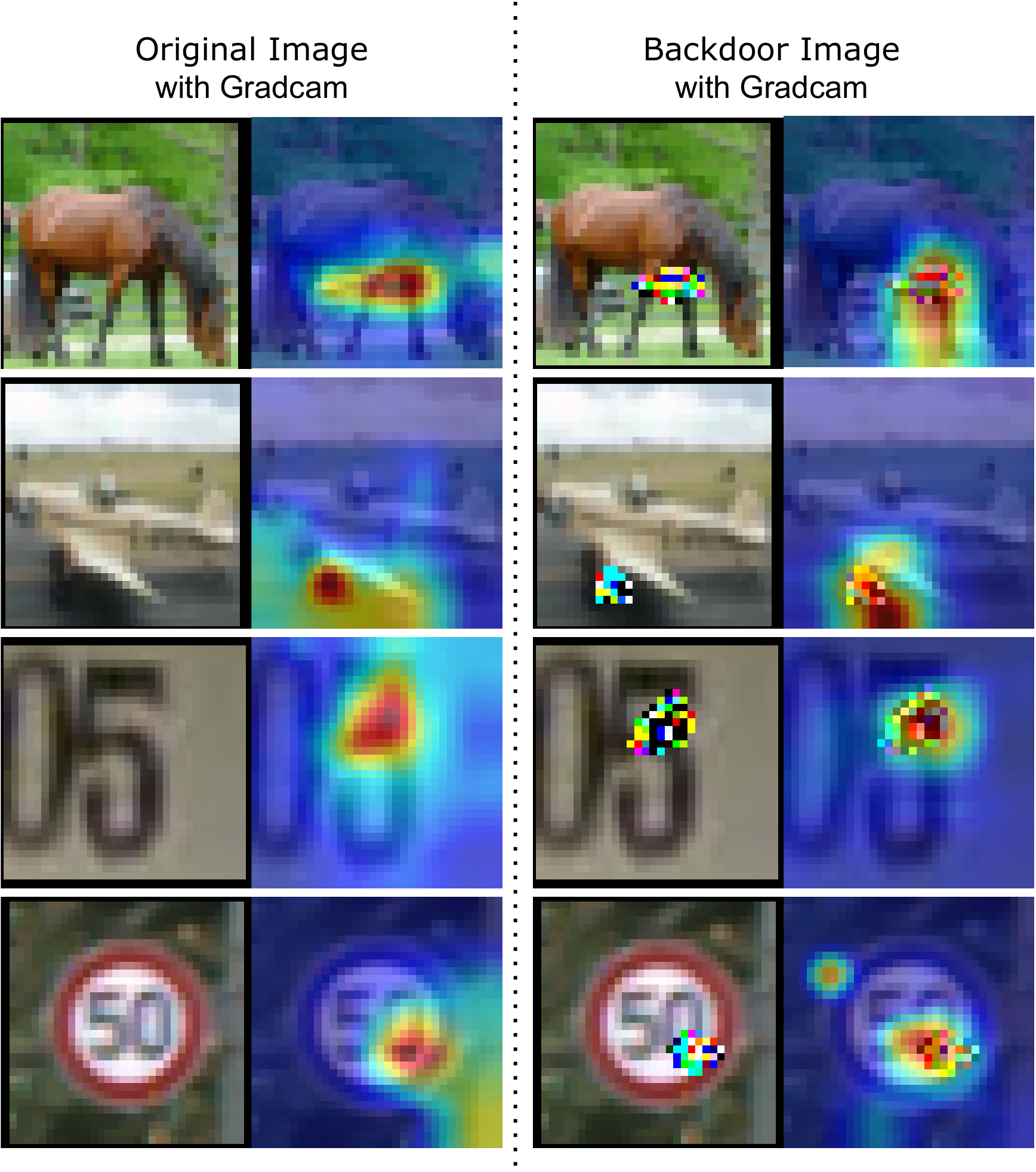}
    \caption{Examples of GradCam maps with and without the trigger}
    \label{fig:gradcamAnalysis}
\end{figure}

In Figure~\ref{fig:gradcamAnalysis}, we present examples of the GradCam results on the original set of images and the respective backdoored version.
As we can see from the heatmaps produced by Gradcam on the original image and the same with the addition of the trigger, the model to classify the image focuses on the same region.
This shows how our trigger successfully substitutes the main features of the original class as the most important portion of the image for the model to classify it.
Another interesting result we can observe is the area of importance detected by the GradCam algorithm.
The different shades of colors presented in Figure~\ref{fig:gradcamAnalysis} represent the importance of those features for the classification of the image. 
The regions with the tones red and yellow are actually the ones that matter the most for the model to perform the classification.
Looking at the examples in Figure~\ref{fig:gradcamAnalysis}, we can see how the produced heatmaps on the original images highlight, of course, the most important region in red but also show how the model looks at regions around in yellow to take a decision for the classification of the image.
The results produced on the backdoored images, instead, show an important region that is much more concentrated around the trigger.
This proves that the combination of the distilled trigger and explainability can confidently drive the target model toward the target class with more confidence compared to the original class. 

To better prove our intuition, we performed an ablation study to test the different components of our approach.
In particular, we tested the components of our solution in three different configurations.
In the first scenario, the trigger is distilled in a fixed position like traditional dynamic solutions.
In the second case, the trigger is still distilled in fixed coordinates, but, differently from the previous scenario, we use GradCam to detect the most important features of the images, and we override them with random noise.
The final scenario, instead, is our full solution.

In Table~\ref{tab:dynamicPatch}, we report the result of our ablation study.
As expected, the dynamic trigger in a fixed position is not capable of producing an effective backdoor attack, confirming what we have already reported in Section~\ref{sub:perfromance_evaluation}.
An interesting result can be seen in the scenario in which we obfuscate the features of the original class. 
As we can see, the success rate of the attack in this scenario improves compared to the one in which we just distill the trigger in a fixed position of the image.
Especially for the SVHN dataset, we can see how the performance of the attack improves three times more after masking the original information of the images.
This result proves how covering the main features of the images and moving the distilled trigger in the right position contribute to deceiving the frozen feature extractor of the transfer model.

\begin{table}
\centering
\caption{Experiments with Dynamic Patch in a fixed position with and without obfuscation of the main features}
\label{tab:dynamicPatch}
\resizebox{0.8\columnwidth}{!}{%
\begin{tabular}{cclclcl}
\hline
Dataset & \multicolumn{2}{c}{\begin{tabular}[c]{@{}c@{}}Dynamic\\ Corner Patch\end{tabular}} & \multicolumn{2}{c}{\begin{tabular}[c]{@{}c@{}}Dynamic\\ Corner Patch\\ With Obfuscation\end{tabular}} & \multicolumn{2}{c}{Our Trigger} \\ \hline
CIFAR10 & \multicolumn{2}{c}{32.3\%}                                                           & \multicolumn{2}{c}{40.0\%}                                                                              & \multicolumn{2}{c}{91.1\%}      \\
CINIC10 & \multicolumn{2}{c}{22.3\%}                                                           & \multicolumn{2}{c}{26.7\%}                                                                              & \multicolumn{2}{c}{73.3\%}      \\
SVHN    & \multicolumn{2}{c}{7.8\%}                                                            & \multicolumn{2}{c}{23.4\%}                                                                              & \multicolumn{2}{c}{72.1\%}      \\
GTSRB   & \multicolumn{2}{c}{18.9\%}                                                           & \multicolumn{2}{c}{33.3\%}                                                                                  & \multicolumn{2}{c}{86.5\%}      \\ \hline
\end{tabular}
}
\end{table}

\subsection{Experiment on Perceptual Similarity Loss}
\label{sub:lpips}

As we introduced in the methodology section (Section~\ref{sub:attackName}), the positioning of our trigger over the main features of the images can alter too much the original data, making it more detectable even by automatic systems that use similarity metrics.
This section is devoted to better understanding how the changes in the loss and trigger initialization described in Section~\ref{sub:blendedTrigger} impact the perceptual similarity between the original and backdoored images using the $LPIPS$ metric presented in the same section.

As we can see in Table~\ref{tab:lpipsValues}, the positioning of the trigger over the original content of the image actually perceptually impacts the most compared to other backdoor attacks that use a static trigger positioned in a peripheral position of the image without impacting the main subject.
With the modified version of the loss, we want to preserve the attack performance, reducing the $LPIPS$ metric.

\begin{table}[h]
\centering
\caption{LPIPS values of the images with different triggers}
\label{tab:lpipsValues}
\resizebox{0.78\columnwidth}{!}{%
\begin{tabular}{ccccc}
\hline
\begin{tabular}[c]{@{}c@{}}Trigger\\ Type\end{tabular} & CIFAR10 & CINIC10 & SVHN   & GTSRB  \\
\hline
Square Trigger                                         & 0.0015  & 0.0007  & 0.0058 & 0.0093 \\
Pattern Trigger                                        & 0.0006  & 0.0002  & 0.0022 & 0.0031 \\
Watermark Trigger                                      & 0.0056  & 0.0020  & 0.0188 & 0.0174 \\
A3FL~\cite{NEURIPS2023_A3FL}                           & 0.0020  & 0.0014  & 0.0106 & 0.0175 \\
FB-FTL (our)                                     & 0.1346  & 0.0618  & 0.2903 & 0.1346 \\
\bottomrule
\end{tabular}
}
\end{table}

In this experiment, we tested the success rate of our approach in normal conditions compared to the more visually conservative version of it.
In particular, we tested the modified version of the original loss with the addition of the $LPIPS$ component combined with random initialization of the trigger and initialization with the original values of the target image.
The results are presented in Table~\ref{tab:lpipsInitialization}.
As we can see, with the more conservative loss, we lower the $LPIPS$ values while preserving most of the performance of the original attack.
Looking at the SVHN datasets, it is interesting to see how the $LPIPS$ version of the loss combined with trigger initialization from the original values of the image performs better than random initialization and even better than the standard version of the attack.
This can be related to the nature of these datasets, as described in Appendix~\ref{sub:datasets}.
In particular, SVHN is characterized by images with backgrounds that can contain features from classes different from the principal one. Adding a trigger too different from the overall image could make the model focus more on the background containing features belonging to categories different from our target class.
A more blended trigger, in this case, allows us to improve even more the performance of our attack.

\begin{table}
\centering
\caption{Backdoor Success Rate (BSR) and $LPIPS$ Value changing the loss and trigger initialization}
\label{tab:lpipsInitialization}
\resizebox{0.8\columnwidth}{!}{%
\begin{tabular}{ccccccc}
\hline
\multirow{2}{*}{Dataset} & \multicolumn{2}{c}{Normal/Noise}                               & \multicolumn{2}{c}{$LPIPS$/Noise}                              & \multicolumn{2}{c}{$LPIPS$/Original}                           \\ \cline{2-7} 
                         & BSR    & \begin{tabular}[c]{@{}c@{}}$LPIPS$\\ Value\end{tabular} & BSR  & \begin{tabular}[c]{@{}c@{}}$LPIPS$\\ Value\end{tabular} & BSR  & \begin{tabular}[c]{@{}c@{}}$LPIPS$\\ Value\end{tabular} \\ \hline
CIFAR10                  & 91.1\% & 0.1346                                                & 84.4\% & 0.0308                                                & 82.2\% & 0.0294                                                \\
CINIC10                  & 73.3\% & 0.0618                                                & 70.0\% & 0.0087                                                & 67.8\% & 0.0076                                                \\
SVHN                     & 72.1\% & 0.2903                                                & 70.0\% & 0.1036                                                & 76.7\% & 0.0919                                                \\
GTSRB                    & 86.5\% & 0.1346                                                & 81.1\% & 0.0375                                                & 78.7\% & 0.0363                                                \\ \hline
\end{tabular}%
}
\end{table}

\begin{figure}[!ht]
    \small
    \centering
    \includegraphics[width=0.25\textwidth]{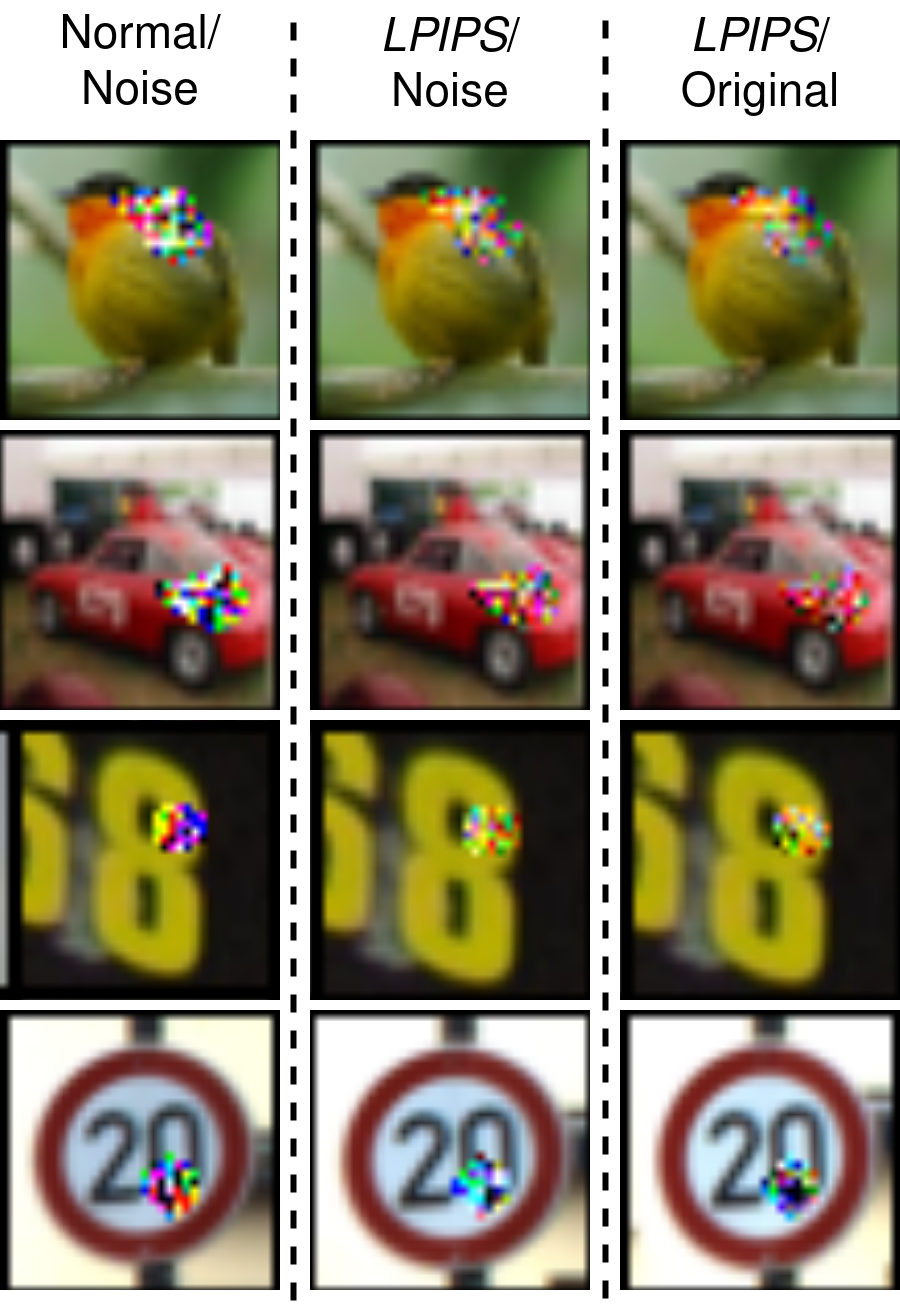}
    \caption{Examples of triggers generated with different loss and trigger initialization}
    \label{fig:LpipsAnalysis}
\end{figure}

In Figure~\ref{fig:LpipsAnalysis}, we show examples of triggers generated with different combinations of losses and initializations.
We can notice that the addition of $LPIPS$ to the basic loss allows the generation of triggers that blend better with the overall image.
The colors of the pixels of the triggers better match the palette and the intensity of the colors of the original image compared to the basic implementation of the dynamic triggers.
Looking at the last column, the initialization of the trigger plays an important part.
As we can see, the generated trigger blends better with the main subject of the image, preserving some of the original pixels that do not need to be updated and resulting in an almost transparent trigger.
The contribution of the initialization strategy is to make the trigger even less noticeable to the human eye compared to other backdoor attacks in which the trigger is easily detectable by a human check.

\subsection{Evaluation on Different Architectures}

\begin{table}[!ht]
\centering
\caption{Evaluation on different architectures}
\label{tab:archEval}
\resizebox{0.6\columnwidth}{!}{%
\begin{tabular}{cccc}
\hline
Dataset & Resnet18 & VGG16  & ConvNet \\ \hline
CIFAR10 & 91.1\%   & 86.7\% & 93.3\%  \\
GTSRB   & 86,5\%   & 85.5\%    & 90.0\%  \\ \hline
\end{tabular}
}
\end{table}

In this experiment, we want to evaluate our approach with additional model architectures to prove the generalizability of our approach.
As mentioned in Section~\ref{sub:experimentalSetup}, we selected two models, namely ConvNet~\cite{gidaris2018dynamic} and VGG16~\cite{simonyan2014very}.
The choice of these two architectures has been made by looking at the architectures studied in the Dataset Condesation paper~\cite{zhao2020dataset}, which inspired our distillation strategy, and the GradCam paper~\cite{selvaraju2017grad}.
For the evaluation of these two architectures, we selected two of the previously presented datasets.
In particular, CIFAR10 as a baseline, and GTSRB since it is the dataset with the highest number of labels.

In Table~\ref{tab:archEval}, we present the success rate of our approach with the two additional architectures.
Our approach, in both cases, preserves most of the accuracy of the considered baseline.
As expected with a deeper network and more parameters like VGG16, the success rate is only partially affected with a decrease of just around $5\%$.

\section{Defenses Evaluation}
\label{sec:Defenses}

In this section, we present the performance of our attack against possible defenses.
Since there are no defenses available for the FTL scenario, we borrowed countermeasures from the other two Federated Learning settings.
First, due to the similarity between the two settings, we start by considering HFL defenses. In particular, we selected two standard ones, Krum and Trimmed Mean, and two more advanced ones, FoolsGold and Flame.
The results on these are presented in Section~\ref{sub:HFLDef}. 

Considering, instead, the VFL scenario, it is difficult to adapt well-established solutions (e.g., Privacy-Preserving Deep Learning or DiscreteSGD) due to the difference between VFL and TFL.
Still, some ideas from a particular family of countermeasures for the Vertical setting can be borrowed for the initial pre-training of the transfer model.
Specifically, solutions from the Label Differential Privacy (LabelDP) family can be exploited to pre-train the transfer model using noisy labels.
In this case, we considered a novel solution named KDk~\cite{arazzi2024kdk} that generates noisy labels using knowledge distillation, and the results are presented in Section~\ref{sub:VFLDef}.

With this study, we want to understand the weaknesses of our attack and develop a working countermeasure designed properly for the Federated Transfer Learning scenario.

\subsection{Horizontal Federated Learning Defenses}
\label{sub:HFLDef}

As we said before, we tested our attack against four different defenses: Trimmed Mean~\cite{yin2018byzantine}, Krum~\cite{blanchard2017machine}, Flame~\cite{nguyen2022flame}, and FoolsGold~\cite{fung2020limitations}.
In the first approach, the server independently averages the gradients according to their position. 
Specifically, the aggregator sorts the gradients in the same position according to their distance from the median. Then, only the first $k$ parameters are considered benign, where $k=n-m$, $n$ is the number of clients, and $m$ is the malicious portion. To be consistent with the original implementation of the defense, we consider the ideal scenario in which the portion of malicious clients is known.
Krum, instead, averages the global model by selecting the best updates between the gradients received from the clients and discarding the outliers that differ significantly from their average.

The Flame defense is composed of two main components: the first filters the clients' updates, and the second adds Adaptive Differential Privacy.
The first component filters malicious clients from the ones with the highest probability of being honest.
This is done by clustering the clients using HDBSCAN over the pairwise cosine similarity distances among the updates and keeping only the cluster that includes at least 50\% clients. 
The second component applies an adaptive differential privacy approach estimating the clipping bound and level of noise, in a way that neutralizes the attack and preserves the original performance of the model.

FoolsGold adjusts the contribution of each client based on the similarity distance of the updates, similar to Flame, and also considers information derived from past iterations.
The approach leverages cosine similarity to measure the distance between updates.
Usually, backdoor attacks alter specific features, which can be identified by measuring the magnitude of model parameters in the output layer of the global model. The malicious updates can be removed or re-weighted.

\begin{figure*}[!ht]
    \centering
    \includegraphics[width=0.75\textwidth]{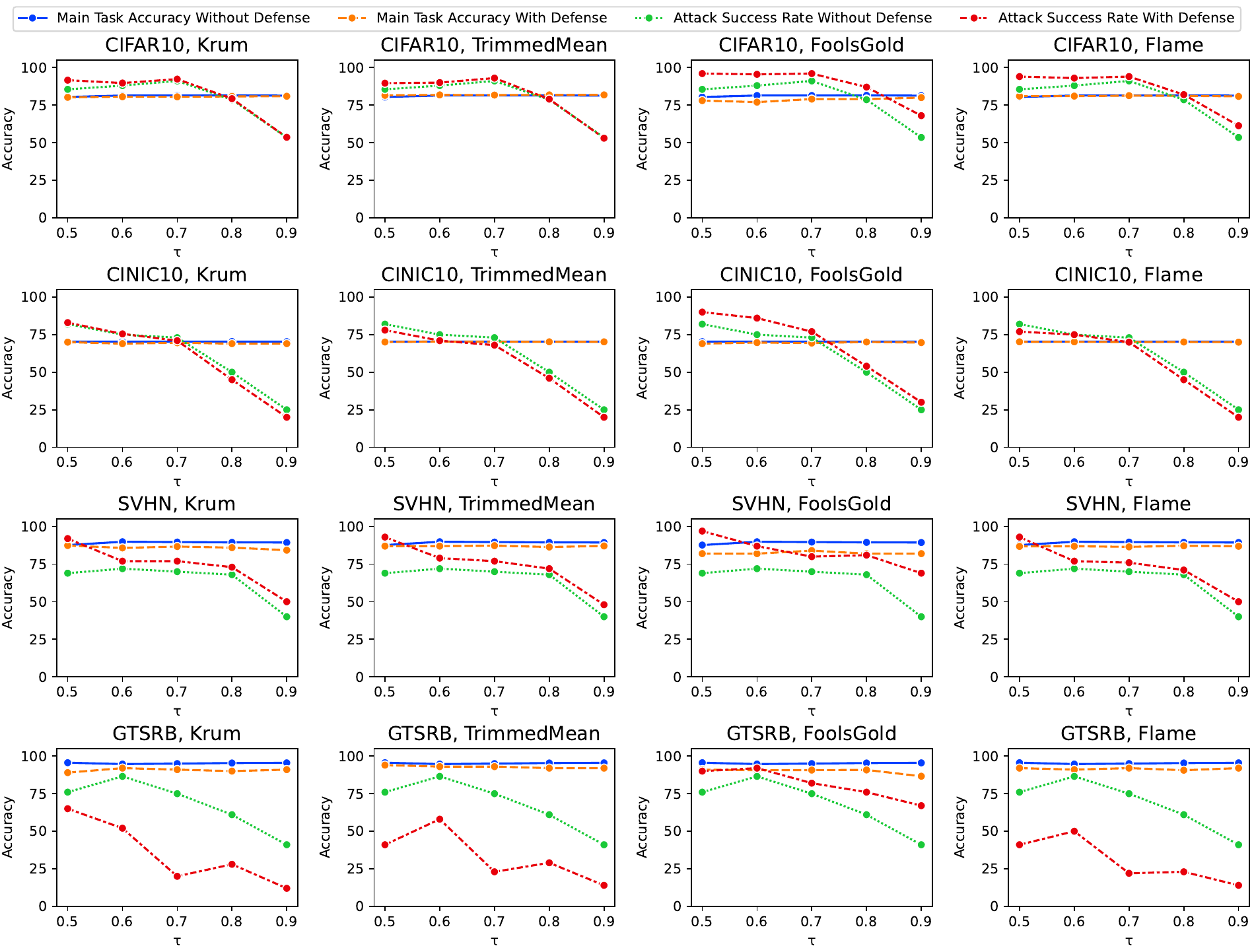}
    \caption{Possible defenses.}
    \label{fig:possibleDefense}
\end{figure*}

In Figure~\ref{fig:possibleDefense}, we present the result of our attack considering different values of $\tau$ against the selected defenses.
Overall, as expected, our attack can pass the filtering of the defenses, preserving the same performance in most cases. This is due to the fact that our triggers are distilled to include the most general features of the target class and override the one of the original class.
As a result, our attack generates updates that are only slightly different from the distribution of the benign ones, making it difficult to detect them.

Interestingly, we noticed two exceptions, one in favor of and one against our solution.
Starting from this last one, we can notice how our attack is penalized against three out of four defenses for the GTSRB dataset. 
As we stated in Appendix~\ref{sub:datasets}, this dataset is characterized by many more categories than the other datasets, $43$ in total, that share many features, resulting in a very small separation between feature distributions across the classes.
The narrower distribution of the updates makes our malicious updates, even with small differences, more easily detectable as outliers by the defenses.
In this scenario, the defenses are capable of mitigating the attack but still are not enough to neutralize scoring, still, an accuracy above $50\%$ with lower values of $\tau$.

The second exception, instead, the one in favor of our solution, is in relation to the results across all the datasets with the employment of FoolsGold as a defense.
As we can see, with all the datasets, our approach archives results in terms of success rate even better compared to the baseline without the defense.
FoolsGold, by design, gives more importance to the features that are beneficial for the accuracy of the final model.
As a result, it will preserve the updates that are closer to the mean of the distribution, penalizing the outliers. 
By definition, our approach aims to distill into the trigger the most general features of the target class, resulting in updates that, on purpose, want to deviate as little as possible from the mean of the benign updates.
To conclude, we demonstrate that defenses intended for Horizontal Federated Learning are effective only in particular cases with datasets with specific characteristics. Moreover, in some situations, they can be beneficial for our attack.

\subsection{Vertical Federated Learning Defense}
\label{sub:VFLDef}

As we have mentioned before, it is challenging to adapt most of the traditional VFL defenses to our scenario. Instead, we can borrow some ideas from LocalDP for the pre-training of the transfer model.
By default, the model is trained with one-hot encoded labels, and it has been proven that this could lead to overfitting problems. To overcome this issue, strategies of label smoothing have been proposed. 
Label smoothing~\cite{szegedy2016rethinking} approaches redistribute the probability of the main class across other secondary classes to instill in the model the knowledge about similarities between classes, making it more robust.
Knowledge Distillation is the perfect candidate to generate a smooth version of the labels.
In practice, Knowledge Distillation uses the output of a pre-trained teacher model to generate smooth labels for each sample.
In this sense, we selected the KDk defense to add a level of uncertainty to the transfer model.

The KDk~\cite{arazzi2024kdk} defense is a novel approach intended to preserve the privacy of the labels in the Vertical Federated scenario.
This solution is based on the concept of $k-anonimity$ between the classes.
In particular, the knowledge distillation is used to select the top-k classes for each image, and then the probability of the main classes is redistributed to the selected top $k$ categories using an $\epsilon$ parameter to tune the re-balancing and the strength of the defense.
In particular, the $\epsilon$ value is subtracted from the probability of the main class, originally set at $1$ in the one-hot setting, and redistributed equally across the $k-1$ secondary labels.
Pushing the $k$ and $\epsilon$ parameters to extreme values will benefit the security of the model at the expense, though, of its performance on the main task.
The tuning of these parameters is important to select the maximum power level without affecting the accuracy of the model.
With this strategy, the features will be learned by the network as less separated between the classes compared to using one-hot encoded labels.
We expect that this additional layer of uncertainty can affect the performance of our attack.

Since our attack is based on distilling into the triggers the general features of the target category, these blurred boundaries between the features of the classes could lead our trigger to also include information from other categories affecting the success rate.
As we said, the selection of the $k$ secondary classes and $\epsilon$ value is important to preserve the accuracy of the model. 
In this experiment, we selected the highest values for both parameters for each dataset that preserve most of the models' accuracy just before they start degrading. 
The selected parameters are reported in Table~\ref{tab:kdkParam}.


\begin{table}[ht]
\centering
\caption{KDk Parameters}
\label{tab:kdkParam}
\resizebox{0.75\columnwidth}{!}{%
\begin{tabular}{ccccc}
\hline
KDk Parameters & CIFAR10 & CINIC10 & SVHN & GTSRB\\
\hline
k value & 3 & 3 & 3 & 5        \\
$\epsilon$ value & 0.45 & 0.4 & 0.3 & 0.6 \\

\hline
\end{tabular}
}
\end{table}

\begin{table}[ht]
\centering
\caption{FB-FTL results against the KDk defense}
\label{tab:kdk}
\resizebox{0.75\columnwidth}{!}{%
\begin{tabular}{ccccc}
\hline
\multirow{2}{*}{Dataset} & \multicolumn{2}{c}{No Defense}                                        & \multicolumn{2}{c}{KDk\cite{arazzi2024kdk}}                                               \\ \cline{2-5} 
                         & \begin{tabular}[c]{@{}c@{}}Main Task\\ Accuracy\end{tabular} & BSR    & \begin{tabular}[c]{@{}c@{}}Main Task\\ Accuracy\end{tabular} & BSR    \\ \hline
CIFAR10                  & 81.3\%                                                       & 91.1\% & 80.1\%                                                       & 84.4\% \\
CINIC10                  & 70.2\%                                                       & 73.3\% & 69.7\%                                                       & 61.1\% \\
SVHN                     & 89.6\%                                                       & 72.1\% & 89.4\%                                                       & 32.2\% \\
GTSRB                    & 94.7\%                                                       & 86.5\% & 94.2\%                                                       & 82.2\% \\ \hline
\end{tabular}
}
\end{table}

In Table~\ref{tab:kdk}, we present the results of our attack against the KDk defense.
Our attack is capable of preserving most of its performance on three of the selected datasets, but at the same time, all of them, even if slightly, are affected by the defense compared to the one for the Horizontal scenario.
What makes our attack resilient is the distillation logic behind the generation of the trigger.
Our strategy tries to distill images by providing examples instead of backpropagating to the information of the target class using just the cross-entropy loss. This makes our solution more robust even if the transfer model has been trained with a level of uncertainty on the separation between the features of the classes.
Similar to what happened with GTSRB against HFL defenses, also in this case, one of the datasets, SVHN, is affected the most by the defense due to its characteristics.
As we said in Appendix~\ref{sub:datasets}, the images of this dataset are characterized by backgrounds that contain information about other classes compared in addition to the principal one. 
Using the distilled soft labels, the KDk defense is training the model to give more importance to the elements in the background mitigating our strategy of overriding the features on the main class.

These results give us insight into how to define a proper defense for our backdoor attack in the Federated Transfer Learning scenario.
In particular, the server should apply an ad-hoc LabelDP strategy for the pre-training of the transfer model instead of focusing on the updates of the clients like the HFL defenses.

\section{Related Work}
\label{sec:related}

Federated Learning was proposed by researchers in Google~\cite{fl-mcmahan}. This work introduced the HFL paradigm, where all the clients use data from the same feature space but may not be of the same sample identity. In VFL, the clients own data that belong to the same samples but do not share the same feature space~\cite{li2020review}. FTL is applied to scenarios in which the clients' datasets are different both in the feature space and the samples identities~\cite{yang2019federated}. 
Given that such scenarios are more general, they are more practical and applicable to real-world applications.
For example, FedHealth~\cite{chen2020fedhealth} uses FTL to support personalized healthcare through wearable devices. To this end, first, the users' devices collaborate to train a shared model in the cloud for human activity recognition with privacy guarantees, and then each participant uses FTL to personalize the model and improve its performance based on the local dataset. 
For the personalization, the authors assumed that in each client, the first part of the downloaded model is frozen, and only the fully connected layers are fine-tuned with the private dataset. We adopted this model in our experiments as the frozen feature extractor that each client has makes traditional backdoor attacks more challenging. The adversary cannot use triggers that are small features~\cite{badnets} and are based on the model's feature extractor, as they can affect only the last layers of the model. As a result, more elaborate strategies need to be designed.

Backdoor attacks in deep learning were first introduced by Gu et al.~\cite{badnets}. In this attack, an adversary alters some training samples to insert a secret functionality into the trained model that is activated during inference. This attack has become very popular and has also been applied to Federated Learning~\cite{bagdasaryan2020backdoor,wang2020attack,xie2019dba,xu2022more,arazzi2023turning}. In~\cite{bagdasaryan2020backdoor}, a model replacement attack was introduced where the adversary attempts to replace the global model with a malicious one that contains a backdoor by amplifying the gradient updates of the adversaries. In~\cite{wang2020attack}, the authors established a theoretical framework to verify that a model is vulnerable to backdoor attacks if it is also vulnerable against evasion attacks in FL.
Xie et al.~\cite{xie2019dba} split the trigger among multiple malicious clients, making the first distributed backdoor attack in the Federated Learning setup. The backdoor can be activated with both the local triggers but also with the global trigger, which is a combination of all the local triggers. Xu et al.~\cite{xu2022more} used this technique to backdoor federated graph neural networks. All these attacks, however, have been applied to HFL, where the adversary has full control of the training of the local model so every layer can be affected. In this work, the adversary cannot alter the feature extractor of the model, making the trigger generation more challenging. To the best of our knowledge, we are the first to explore backdoor attacks in this scenario.

\section{Conclusions and Future Work}
 \label{sec:conclusions}

In this paper, we describe, to the best of our knowledge, the first focused backdoor attack specifically designed for the Federated Transfer Learning Scenario.
Our attack, called FB-FTL, aims to overcome the challenges imposed by the scenario. 
In this context, a feature learning step is carried out by the server before the actual federated learning, which then aims at collaboratively train only the classification layers by keeping frozen the feature extractor ones. 
Due to this characteristic, existing backdoor attacks, especially the static ones that introduce unknown features to the feature extractor, are ineffective in this scenario.
Moreover, traditionally, backdoor attacks introduce a trigger in the original data without applying any strategy to minimize its impact on them. For instance, in the case of image data, the position of the trigger may strongly influence the visualization outcome.

To design our attack, we started by analyzing the FTL framework to understand its vulnerabilities and to exploit them.
The main intuition behind our approach is to maximize the success of our attack by identifying the optimal spot in the input data to position the trigger.
To do this, we leverage an XAI strategy to identify the most important regions of the images for their classification. The use of explainability to locate the trigger over features detectable by the feature extractor component of the FTL model brought to the definition of the first focused backdoor attack.
Another important aspect of our attack is related to the way in which the trigger is generated.
For this purpose, we borrowed some ideas from the dataset distillation field. In particular, the attack distills the main features of the target class into the trigger. 
Combined with the focusing strategy, it overrides the information of the original class with the desired one.
In our experimental campaign, we analyzed the different components of our approach, showing the importance of distilling the target class features and correctly positioning the trigger in the input data.
We demonstrated how our approach does not depend on the percentage of malicious clients in the process and is not heavily affected by the total number of clients.
Since no ad-hoc defenses are available for the considered scenario, we evaluated our attack against Horizontal and Vertical Federated Learning countermeasures. 
As we discovered, no defense is completely effective in mitigating our approach for all the datasets.
In this sense, the most promising approach comes from the Vertical Federated family, named KDk, and makes use of Lable Differential Privacy (LabelDP) to add uncertainty between the classes in the pre-trained model.
In this direction, as future work, further exploration can be conducted to define a more effective countermeasure, still based on LabelDP, against our attack.

\appendix

\section{More Examples}
\label{sec:moreExamples}

\begin{figure*}[!ht]
    \centering
    \includegraphics[width=0.6\textwidth]{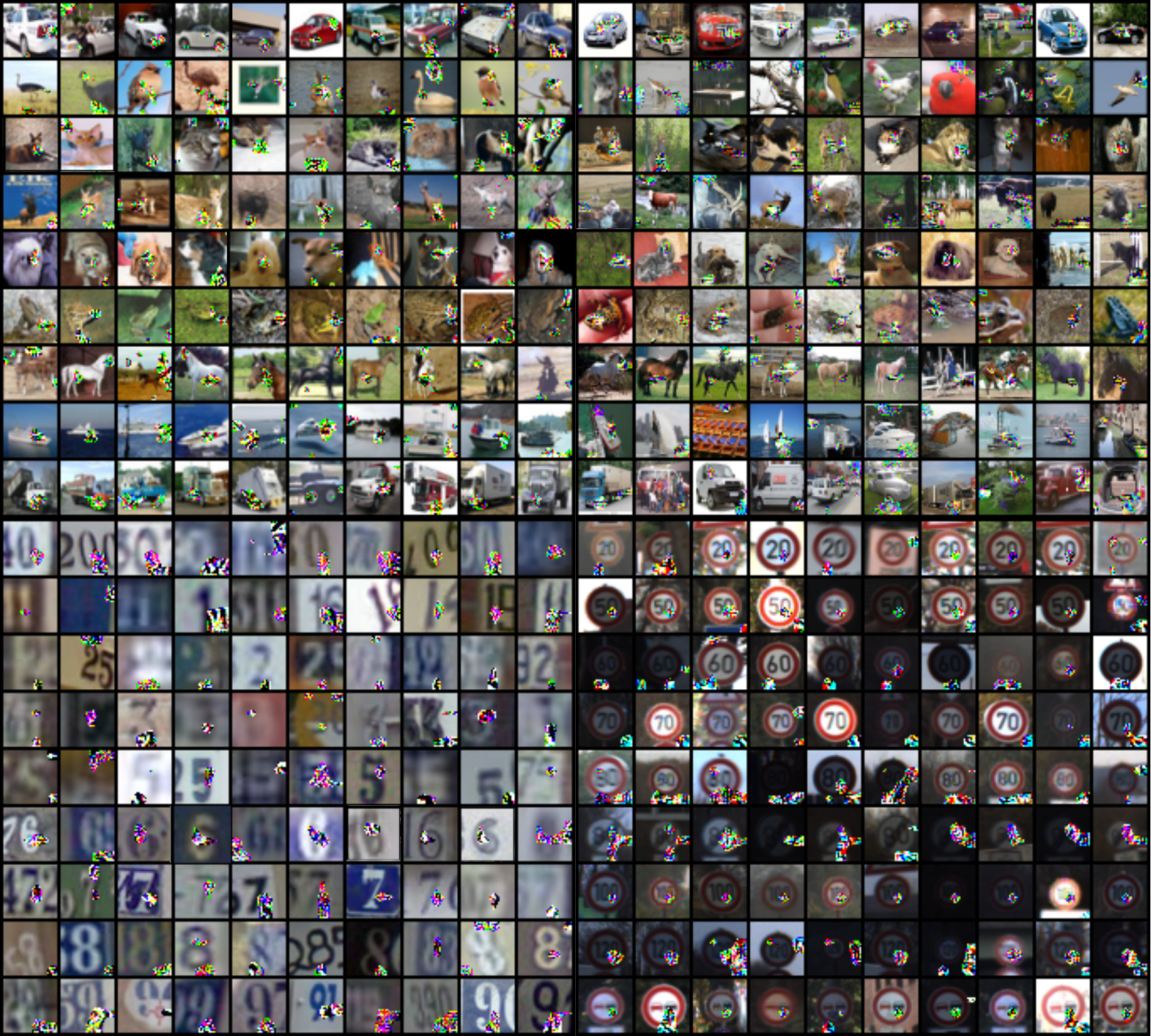}
    \caption{More examples of triggers generated with the basic implementation.}
    \label{fig:moreExamplNormal}
\end{figure*}

\begin{figure*}[!ht]
    \centering
    \includegraphics[width=0.6\textwidth]{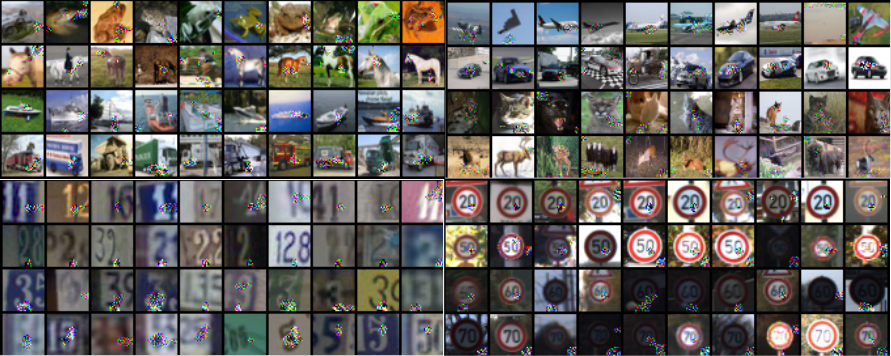}
    \caption{More examples of triggers generated with the contribution of $LPIPS$ in the loss.}
    \label{fig:moreExamplLpips}
\end{figure*}

\begin{figure*}[!ht]
    \centering
    \includegraphics[width=0.6\textwidth]{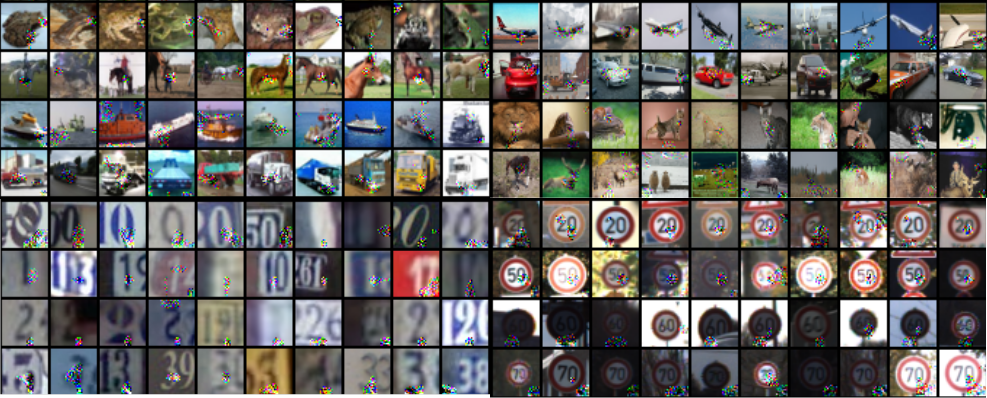}
    \caption{More examples of triggers generated with the contribution of $LPIPS$ in the loss and the initialization of the trigger with the original values of the image.}
    \label{fig:moreExamplOriginal}
\end{figure*}

In this section, we show more examples of triggers generated by our attack.
In Figure~\ref{fig:moreExamplNormal}, examples of the basic implementation of our attack are shown.
In Figure~\ref{fig:moreExamplLpips} and \ref{fig:moreExamplOriginal}, instead, we show examples of the $LPIPS$ contribution in the loss and the initialization of the trigger with the original values of the images.

\section{Datasets}
\label{sub:datasets}

\begin{table}[ht]
\centering
\caption{Statistics of the considered datasets}
\label{tab:datasetStats}
\begin{tabular}{cccc}
\hline
Dataset & \begin{tabular}[c]{@{}c@{}}Train\\ Set Size\end{tabular} & \begin{tabular}[c]{@{}c@{}}Test\\ Set Size\end{tabular} & \begin{tabular}[c]{@{}c@{}}Number of\\ Classes\end{tabular} \\ \hline
CIFAR10 & 50,000                                                   & 10,000                                                   & 10                                                          \\
CINIC10 & 180,000                                                  & 90,000                                                   & 10                                                          \\
SVHN    & 73,257                                                   & 26,032                                                   & 10                                                          \\
GTSRB   & 39,209                                                   & 12,630                                                   & 43                                                          \\ \hline
\end{tabular}
\end{table}

To conduct the experimental campaign, we selected four of the most common benchmark datasets: CIFAR10~\cite{krizhevsky2009learning}, CINIC10~\cite{darlow2018cinic}, SVHN~\cite{netzer2011reading}, and GTSRB~\cite{Stallkamp2012}.
The first two are datasets that include classes of different animals or objects. Both share the same categories. The main difference is that CINIC10 is 4.5 times larger the size of CIFAR10. This difference in size allows us to prove that our approach is still effective even with larger datasets, proving our intuition that the trigger is distilled, preserving the most general and meaningful features of the target class.
The SVHN dataset, instead, contains images of houses' civic numbers with categories from 0 to 9. One of the main characteristics of this dataset is the presence of additional numbers in the background that do not belong to the assigned category. We selected this dataset to see if the features distilled in our trigger are strong enough to be preferred to the other number in the image by the model.
The last dataset, the German Traffic Sign Recognition Benchmark (GTSRB), contains forty-three different categories of traffic signs. Since the traffic signs have many features in common between the categories (e.g., shape and color), we included this dataset to test whether our approach is capable of overriding the right features to control the model and predict the target class. 
Table~\ref{tab:datasetStats} reports statistics for the considered datasets.

\end{document}